\documentclass[a4paper,fleqn]{cas-sc}

\usepackage[authoryear]{natbib}
\usepackage{amsmath}
\usepackage{graphicx}
\usepackage{booktabs}
\usepackage[ruled,vlined]{algorithm2e}

\usepackage{amssymb}
\usepackage{adjustbox}
\usepackage{subcaption}
\usepackage{tabularx}
\usepackage{array}
\usepackage{multirow}
\usepackage{pifont}
\usepackage[utf8]{inputenc}
\usepackage{url}
\usepackage{mathtools}
\usepackage{xcolor}

\ExplSyntaxOn
\cs_set:Npn \__first_footerline:
{
  \group_begin:
  \small
  \sffamily
  \ifnum\theblind>0\relax
  \else
    \__short_authors:
  \fi
  \group_end:
}
\ExplSyntaxOff

\begin{document}
\let\WriteBookmarks\relax
\def\floatpagepagefraction{1}
\def\textpagefraction{.001}

% Short title
\shorttitle{\text{Knowledge-Guided Distributional Prompt Learning with Visual Hierarchy for Zero-Shot Anomaly Detection}}

% Main title of the paper
\title[mode = title]{Knowledge-Guided Distributional Prompt Learning with Visual Hierarchy for Zero-Shot Anomaly Detection}

\shortauthors{J. Ham et~al.}

\author{Jiyul Ham}
\fnmark[1]
\ead{jiyulham@korea.ac.kr}

\author{Seongho Yoon}
\fnmark[1]
\ead{songhow53@korea.ac.kr}

\author{Yonggon Jung}
\ead{yg4458@korea.ac.kr}

\author{Sangjae Lee}
\ead{kr_sjsjsj@korea.ac.kr}

\author{Jun-Geol Baek}
\cormark[1]
\ead{jungeol@korea.ac.kr}

\affiliation{
    organization={Department of Industrial and Management Engineering, Korea University},
    addressline={145 Anam-ro, Seongbuk-gu},
    city={Seoul},
    postcode={02841},
    country={Republic of Korea}
}

\cortext[cor1]{Corresponding author}
\fntext[fn1]{These authors contributed equally to this work.}

% Here goes the abstract
\begin{abstract}
Zero-shot anomaly detection (ZSAD) performs generalized anomaly detection on previously
unseen classes by using auxiliary images. ZSAD plays a vital role as a practical anomaly
detection approach in various fields, such as industrial visual inspection and medical imaging, where annotation costs are high and rapid decision-making is essential. Recent vision-language-model-based approaches have substantially improved ZSAD by aligning textual representations with visual features, while more recent distributional prompting methods further model prompt variability to enhance generalization to unseen categories. However, existing methods do not explicitly combine normality-referenced distributional modeling with hierarchy-specific prompt specialization, limiting their ability to represent anomalous semantics relative to normality while simultaneously addressing image-level detection and fine-grained localization. To address this limitation, we propose Hierarchical Knowledge-Guided Distributional Prompt Learning (HiDist), a normality-referenced hierarchical distributional prompt framework for ZSAD. HiDist models each prompt token using a diagonal Gaussian distribution and generates stochastic prompt variations through Monte Carlo sampling. Rather than learning normal and anomalous distributions independently, anomalous prompt distributions are parameterized as deviations from learned normal references, explicitly encoding their statistical dependency. The resulting distributions are specialized into global and local representations for image-level detection and pixel-level localization, respectively. An adapted global visual representation then guides hierarchical alignment, together with complementary regularization objectives. Experiments on seven industrial and medical datasets demonstrate strong cross-dataset performance of HiDist across both anomaly detection and localization tasks. Extensive ablation and representation analyses further support the effectiveness of normality-referenced distributional modeling and hierarchy-specific representation learning.
\end{abstract}

% Keywords
% Each keyword is separated by \sep

\maketitle

%% main text

% Introduction
\section{Introduction}
Anomaly detection has become critical in fields requiring rapid identification of irregular patterns~\citep{chandola2009anomaly, zhu2024anomaly}, such as industrial quality control~\citep{tao2022deep} and clinical decision support~\citep{fernando2021deep}. In industrial domains, early detection of defective products is essential for cost reduction and quality assurance, whereas in the medical domain, identifying abnormal regions during diagnosis is vital to assist healthcare professionals in treatment planning and decision-making. Accordingly, achieving high performance and strong generalization capability has become crucial for anomaly detection~\citep{pang2021deep}.

Visual anomaly detection is recognized as an out-of-distribution problem because normal samples are relatively easy to obtain, whereas abnormal samples are extremely sparse and costly to annotate~\citep{ruff2021unifying, tao2022deep}. Owing to these constraints, unsupervised anomaly detection (UAD) has become the dominant approach~\citep{liu2024deep, roth2022towards, zavrtanik2021draem, zavrtanik2021reconstruction}.
Although these methods have demonstrated promising results, they also reveal practical limitations and data-related constraints for real-world industrial deployment. From a practical perspective, most existing models are class-specific, creating challenges for developing, maintaining, and managing separate models across large-scale manufacturing environments with numerous product lines and processes~\citep{salehi2021unified}. Meanwhile, from a data perspective, there are issues of performance degradation because of data scarcity in new environments and distribution shift in new domains~\citep{zhang2023industrial}. Given these practical constraints, developing a universal zero-shot anomaly detection (ZSAD) framework becomes imperative—one that operates across diverse categories without necessitating class-specific model training.

Vision–language models (VLMs) have become a cornerstone of modern ZSAD. In particular, contrastive language–image pretraining (CLIP)~\citep{radford2021learning} demonstrated that joint training on massive image–text pairs can align visual and textual information, enabling impressive zero-shot classification and retrieval without additional fine-tuning.
To improve the contrastive learning objective of CLIP, SigLIP~\citep{zhai2023sigmoid} introduced a sigmoid loss, while SigLIP 2~\citep{tschannen2025siglip} enhanced multilingual understanding and fairness through multilingual captioning and debiasing. These advances in VLMs serve as a foundation for ZSAD research.

\begin{figure}[pos=htbp]
    \centering
    \adjustbox{clip=False}{
        \includegraphics[width=0.97\textwidth]{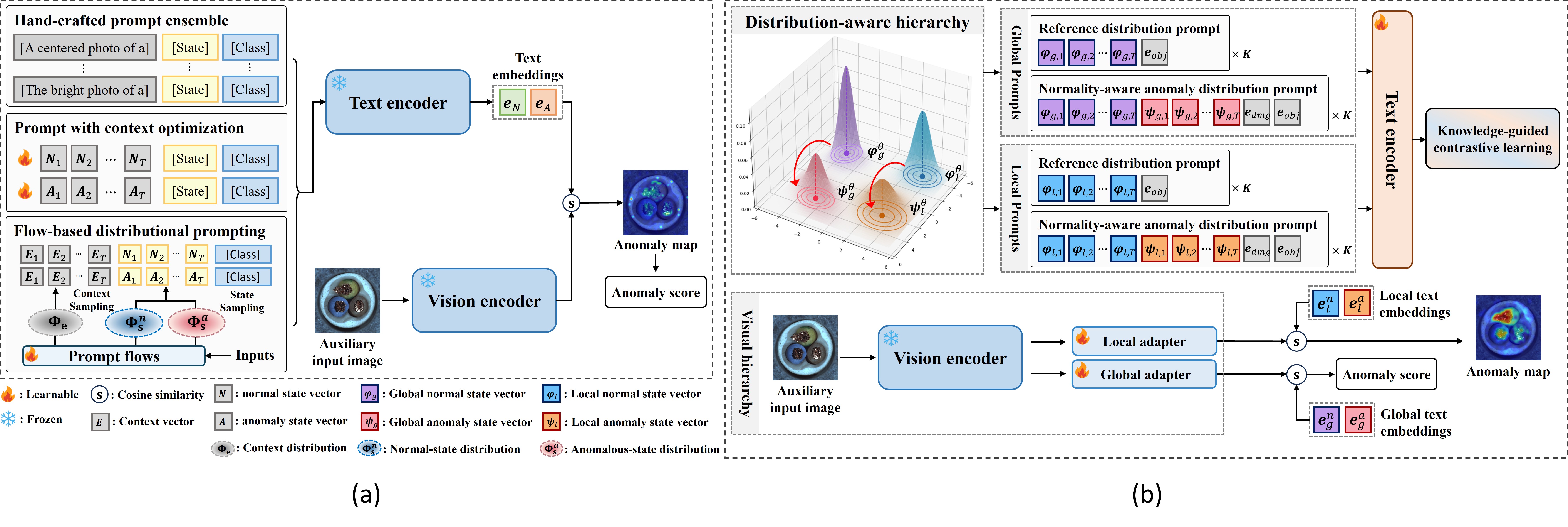}
        }
    \caption{
        Comparison of prompt-learning paradigms for zero-shot anomaly detection (ZSAD) and the proposed HiDist.
        (a) Evolution of prompt design from hand-crafted prompt ensembles and context optimization to flow-based distributional prompting.
        (b) HiDist structures normality-aware distributional prompts within global--local visual hierarchies and aligns them through knowledge-guided contrastive learning.
        }
    \label{fig:prompt design}
\end{figure}

Recent ZSAD research has advanced primarily through text-prompt design and visual-representation enhancement. As illustrated in Fig.~\ref{fig:prompt design}(a), prompt-design strategies have evolved from hand-crafted prompt ensembles to learnable context optimization and, more recently, distributional prompting. APRIL-GAN~\citep{chen2023april} and WinCLIP~\citep{jeong2023winclip} combine fixed templates with multiple words describing normal and anomalous states, whereas AnomalyCLIP~\citep{zhou2024anomalyclip} and AdaCLIP~\citep{cao2024adaclip} replace static contexts with learnable vectors or image-conditioned prompts. Bayes-PFL~\citep{qu2025bayesian} further extends this progression by introducing flow-based probabilistic prompting to generate diverse prompt variants beyond deterministic point estimates. Despite this advance, this line of work primarily emphasizes prompt-space flexibility and coverage, while the anomaly-specific structure of the learned distributions remains less explicitly modeled. In particular, token-specific variability and the explicit relationship between normal and anomalous states remain insufficiently structured, while prompt distributions are not explicitly specialized for the distinct semantic granularities required by image-level detection and pixel-level localization. These observations motivate a more structured probabilistic prompt space that models variability at individual token positions, represents abnormality relative to normality, and distinguishes global from local anomaly semantics.

% Figure 1

In this study, we propose Hierarchical Knowledge-Guided Distributional Prompt Learning (HiDist), a normality-referenced hierarchical distributional prompt framework for ZSAD. As illustrated in Fig.~\ref{fig:prompt design}(b), rather than sampling shared latent offsets that modify deterministic prompt segments, HiDist directly parameterizes each learnable prompt token as an image-agnostic Gaussian distribution with learnable mean and diagonal scale parameters. This token-wise formulation enables individual token positions to learn distinct semantic centers and variability within a prompt sequence. Rather than prioritizing greater distributional flexibility through nonlinear flow transformations, HiDist adopts explicit Gaussian parameterization as a structural inductive bias, providing explicit token-level distribution parameters that can be directly related across normal and anomalous states. Accordingly, each anomalous token distribution is formulated through learnable parameter deviations from its corresponding normal reference distribution, explicitly encoding how anomalous semantics depart from normality. HiDist further organizes these normality-referenced token distributions into separate global and local branches specialized for image-level anomaly detection and pixel-level localization, respectively. An adapted global visual representation provides shared semantic knowledge for aligning the resulting global and local text prototypes, preserving their level-specific roles while maintaining consistency across the visual hierarchy. Finally, prompt samples drawn from each distribution are encoded and averaged to construct normal and anomalous text prototypes at both levels.

The main contributions of this study are as follows: 
\begin{itemize}

\item We introduce a fine-grained distributional prompt formulation that directly parameterizes each learnable prompt token as an image-agnostic Gaussian distribution. Unlike probabilistic prompting based on shared stochastic modifications to deterministic prompt tokens, the proposed token-wise formulation assigns distinct statistical parameters to individual token positions, enabling token-specific semantic centers and variability to be explicitly represented.

\item We propose normality-aware anomaly prompt modeling that exploits the explicit Gaussian parameterization to structure the relationship between normal and anomalous states at the distribution-parameter level. Specifically, each anomalous token distribution is defined through learnable deviations from its corresponding normal reference distribution, representing abnormality as a relative departure from normality rather than as an independently modeled state.

\item We develop a hierarchical distributional prompt design that organizes the normality-referenced distributions into separate global and local branches for image-level anomaly detection and pixel-level localization, respectively. An adapted global visual representation further serves as shared semantic knowledge to align the resulting text prototypes through knowledge-guided contrastive learning, preserving level-specific specialization while maintaining cross-hierarchical consistency.

\item Extensive experiments across industrial and medical anomaly detection datasets validate the effectiveness and generalizability of HiDist, with comprehensive ablation and analysis demonstrating the complementary benefits of its normality-aware distributional modeling and hierarchical representation design.

\end{itemize}

The structure of this paper is outlined as follows. In Section 2, we review related work, and Section 3 introduces the HiDist framework. Sections 4 and 5 present the experimental results and their analysis on industrial and medical datasets. Finally, Section 6 provides the conclusion.

\section{Related work}
\subsection{Vision-language models and prompt learning}
VLMs are designed to bridge visual and textual modalities~\citep{ghosh2024exploring}. In particular, CLIP~\citep{radford2021learning} leverages massive-scale paired data and contrastive learning to construct a joint representation space for both modalities.
To overcome batch size constraints and computational inefficiency in contrastive learning, SigLIP~\citep{zhai2023sigmoid} introduced a sigmoid-based binary matching objective that frames image-text matching as binary classification. SigLIP 2~\citep{tschannen2025siglip} enhances domain generalization by integrating multilingual captioning and debiasing techniques to address multilingual performance and bias issues.
However, zero-shot use of VLMs primarily relies on static prompt templates such as ``\texttt{A photo of a [Class]}'', which limits text representations to class semantics, which constrains their ability to generate task-tailored representations. Consequently, prompt learning has become a prominent approach to enhance the adaptability of VLMs across various tasks. CoOp~\citep{zhou2022learning} proposed optimizing task-specific prompts with learnable prompt vectors to address the limitations of static text templates. CoCoOp~\citep{zhou2022conditional} introduced dynamic prompt generation conditioned on image features to mitigate dataset-specific overfitting. These advances offer new possibilities for semantic alignment of unseen anomaly patterns in ZSAD.

Beyond deterministic prompt optimization, recent studies have further explored probabilistic prompt representations to capture semantic diversity that cannot be sufficiently represented by a single optimized prompt. ProDA~\citep{lu2022prompt} models prompt representations as a distribution to improve generalization across visual domains, while Bayesian Prompt Learning~\citep{derakhshani2023bayesian} formulates prompt optimization from a Bayesian perspective and regularizes the prompt space through probabilistic inference. Probabilistic Prompt Learning~\citep{kwon2023probabilistic} further demonstrates the benefit of sampled textual representations for dense prediction. These studies suggest that modeling prompts as distributions can provide richer semantic coverage than deterministic point estimates.

\subsection{Zero-shot anomaly detection}
ZSAD aims to identify anomalies across novel categories and datasets through learning from auxiliary data~\citep{jeong2023winclip, chen2023april, zhou2024anomalyclip}. Recent studies advanced ZSAD by focusing on text prompt design and visual-feature enhancement.

\textbf{Text prompt design.} Early VLM-based ZSAD methods mainly relied on hand-crafted textual descriptions of normal and anomalous states. WinCLIP~\citep{jeong2023winclip} designed two
state prompts for normal and anomalous states, and proposed compositional prompt ensembles combining context and state from hand-crafted prompt templates. Subsequent approaches shifted toward learnable representations. AnomalyCLIP~\citep{zhou2024anomalyclip} introduces object-agnostic learnable prompts that focus on generic normal and abnormal semantics, whereas AdaCLIP~\citep{cao2024adaclip} combines static and image-conditioned dynamic prompts to adapt textual representations to individual inputs.

More recently, Bayes-PFL~\citep{qu2025bayesian} extends ZSAD prompt learning from deterministic representations to probabilistic prompt spaces. It learns flow-based distributions for image-specific context and image-agnostic state representations, whose samples are added to the corresponding tokens of multiple learnable prompt banks to generate diverse prompt variants. This formulation substantially improves the flexibility and coverage of the prompt space. Nevertheless, each sampled context or state representation is shared across multiple token positions, rather than explicitly assigning distinct distributional parameters to individual prompt tokens. Moreover, normal and anomalous state distributions are generated separately without an explicit parameter-level reference relationship.

\textbf{Visual feature enhancement.}
To improve fine-grained anomaly perception, WinCLIP~\citep{jeong2023winclip} aggregates spatial information using multi-scale windows, whereas APRIL-GAN~\citep{chen2023april} exploits intermediate patch features from multiple encoder layers. AnomalyCLIP~\citep{zhou2024anomalyclip} modifies the attention mechanism to enhance local visual semantics, and AdaCLIP~\citep{cao2024adaclip} further adapts visual representations through learnable projection modules. Despite these advances, image-level anomaly detection and pixel-level localization require different semantic granularities: the former relies more strongly on global structural information, whereas the latter requires fine-grained local evidence. Existing methods generally do not explicitly associate these visual levels with separately specialized textual representations, leaving the distinct semantic requirements of image-level detection and pixel-level localization insufficiently differentiated.

Taken together, these studies highlight two complementary challenges in current ZSAD: structuring normal and anomalous prompt representations beyond flexible prompt-space coverage, and aligning textual representations with the distinct global and local semantics required by detection and localization. HiDist addresses these challenges through normality-referenced token-wise distributional prompting and hierarchical global-local representation learning.

% Figure 2
\begin{figure}[pos=htbp]
    \centering
    \adjustbox{clip=False}{
        \includegraphics[width=0.97\textwidth]{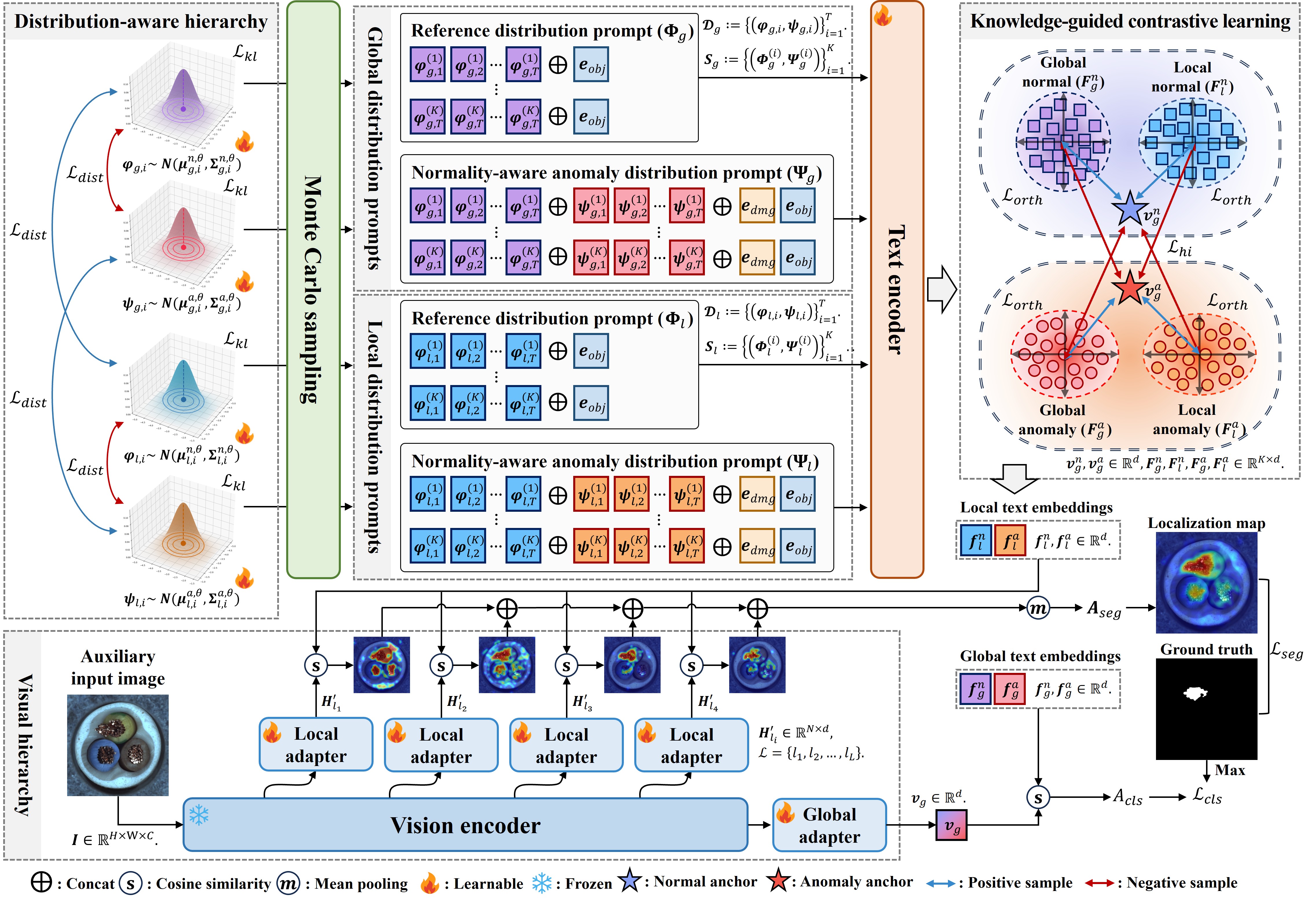}
    }
    \vspace{-1mm}
    \caption{
        Overview of the proposed HiDist framework.
        Given an auxiliary training image, the vision encoder extracts global and multi-scale local visual representations, which are further adapted through the corresponding visual adapters.
        In parallel, token-wise distributional prompts are sampled from normal reference distributions and their corresponding anomalous distributions at the global and local levels, and are encoded into semantic prototypes.
        Knowledge-guided contrastive learning aligns the hierarchical text prototypes with the adapted global visual representation $\boldsymbol{v}_g$, which serves as a shared knowledge anchor.
        }
    \label{fig: HiDist}
\end{figure}

\section{Methodology}

\subsection{Preliminaries: ZSAD with SigLIP}

The vision and text encoders of SigLIP are denoted as $\mathcal{E}_v(\cdot)$ and $\mathcal{E}_t(\cdot)$, respectively. The standard SigLIP prompt template takes the structure ``\texttt{A photo of a [class]}'', where ``\texttt{class}'' refers to a specific object class. Images $\boldsymbol{I}\in \mathbb{R}^{H\times W\times C}$ and prompts $\boldsymbol{P} \in \mathbb{R}^{T \times d}$ are mapped into a joint embedding space:
\begin{equation}
\boldsymbol{v} = \mathcal{E}_v(\boldsymbol{I}), \quad
\boldsymbol{f} = \mathcal{E}_t(\boldsymbol{P}).
\end{equation}
Here, $H$, $W$, and $C$ specify the image height, width, and channel dimensions, respectively; $T$ denotes the number of prompt tokens; and $d$ is the embedding dimension. The mapped features $\boldsymbol{v} \in \mathbb{R}^d$ and $\boldsymbol{f} \in \mathbb{R}^d$ represent visual and textual embeddings, respectively. For anomaly detection, we construct a normal prompt $\boldsymbol{P}^n$ and an anomaly prompt $\boldsymbol{P}^a$, and obtain their corresponding text features $\boldsymbol{f}^n$ and $\boldsymbol{f}^a$.

The image-level anomaly score is calculated as follows:
\begin{equation}
\begin{aligned}
A_{cls}(\boldsymbol{v}, \boldsymbol{f}^a)
&=
\sigma(\widetilde{sim}(\boldsymbol{v}, \boldsymbol{f}^a)/\tau), \\
\widetilde{sim}(\boldsymbol{x}, \boldsymbol{y})
&=
sim\left(
\frac{\boldsymbol{x}}{\|\boldsymbol{x}\|_2},
\frac{\boldsymbol{y}}{\|\boldsymbol{y}\|_2}
\right),
\end{aligned}
\label{anomaly_cls}
\end{equation}
where $\sigma(\cdot)$ denotes the sigmoid function, $\|\cdot\|_2$ denotes the $l_2$ norm, $sim(\cdot,\cdot)$ calculates cosine similarity, and $\tau$ controls the temperature scaling.

To enable pixel-wise localization, we extract the patch-level representation $\boldsymbol{H}_{l_i} \in \mathbb{R}^{N \times d}$ from the intermediate layers $\mathcal{L} = \{l_1, l_2, \ldots, l_L\}$ of $\mathcal{E}_v$, with $N$ denoting the total patch count. The anomaly score map for each layer is computed as follows:
\begin{equation}
\boldsymbol{A}_{seg}^{(i)}(\boldsymbol{H}_{l_i}, \boldsymbol{f})
=
\frac{
\exp(\widetilde{sim}(\boldsymbol{H}_{l_i}, \boldsymbol{f}^a)/\tau)
}{
\exp(\widetilde{sim}(\boldsymbol{H}_{l_i}, \boldsymbol{f}^n)/\tau)
+
\exp(\widetilde{sim}(\boldsymbol{H}_{l_i}, \boldsymbol{f}^a)/\tau)
}.
\label{anomaly_map}
\end{equation}
During inference, the anomaly map is computed by averaging the scores across all layers and applying a Gaussian filter:
\begin{equation}
\boldsymbol{A}_{seg}
=
\mathcal{G}_\sigma
\left(
\frac{1}{L}
\sum_{i=1}^{L}
\boldsymbol{A}_{seg}^{(i)}
(\boldsymbol{H}_{l_i}, \boldsymbol{f})
\right).
\label{agg_anomaly_map}
\end{equation}
Here, $\mathcal{G}_\sigma(\cdot)$ is a Gaussian filter with standard deviation $\sigma$.

\subsection{Overall workflow}

Fig.~\ref{fig: HiDist} illustrates the distribution-aware hierarchy of HiDist through four key components. (1) Distribution-aware prompt modeling represents each prompt token as a diagonal multivariate Gaussian and explores the textual embedding space via Monte Carlo sampling. (2) Normality-aware anomaly prompt modeling constructs anomaly prompts by incorporating the statistical characteristics of normal distributions. (3) A visual hierarchical design with knowledge-guided contrastive learning models distributions at the global and local levels to capture the visual hierarchy specialized through visual adapters. (4) A comprehensive learning strategy integrates multiple training objectives, including distributional separation and sample diversity. Finally, ZSAD is performed using the learned global and local normal/anomalous text prototypes.

\subsection{Distribution-aware prompt modeling}

\textbf{Monte Carlo sampling for prompt generation.}
We propose distribution-aware prompt modeling that represents prompts as probability distributions and enriches the textual embedding space by capturing distributional variability through Monte Carlo sampling. Given a prompt of length $T$ as $\boldsymbol{P} = [\boldsymbol{p}_1, \boldsymbol{p}_2, \ldots, \boldsymbol{p}_T]$, we model each token $\boldsymbol{p}_i \in \mathbb{R}^d$ as an independent diagonal multivariate Gaussian distribution:
\begin{equation}
\begin{aligned}
\boldsymbol{p}_i
&\sim
\mathcal{N}(
\boldsymbol{\mu}_i^{\theta},
\boldsymbol{\Sigma}_i^{\theta}),
\quad i=1,2,\ldots,T,\\
\boldsymbol{\Sigma}_i^{\theta}
&=
\operatorname{Diag}
\left(
(\boldsymbol{\sigma}_i^{\theta})^2
\right),\\
\boldsymbol{\mu}_i^{\theta}
&\in\mathbb{R}^{d},
\quad
\boldsymbol{\sigma}_i^{\theta}
\in\mathbb{R}_{+}^{d}.
\end{aligned}
\end{equation}
Here, $\boldsymbol{\mu}_i^{\theta}$ and $\boldsymbol{\sigma}_i^{\theta}$ denote the learnable mean and per-dimension scale parameters of the $i$th token, respectively. Thus, $\boldsymbol{\Sigma}_i^{\theta}$ is a diagonal covariance matrix, and each token position has its own distributional parameters.

To generate diverse prompts, we sample $K$ prompt instances using the reparameterization trick~\citep{kingma2013auto}:
\begin{equation}
\boldsymbol{p}_i^{(k)}
=
\boldsymbol{\mu}_i^{\theta}
+
\boldsymbol{\sigma}_i^{\theta}
\odot
\boldsymbol{\epsilon}_i^{(k)},
\quad
\boldsymbol{\epsilon}_i^{(k)}
\sim
\mathcal{N}(\mathbf{0},\mathbf{I}),
\label{reparametric}
\end{equation}
where $\odot$ denotes element-wise multiplication and $\boldsymbol{p}_i^{(k)}$ represents the $i$th token of the $k$th sampled prompt. The $k$th prompt is then constructed as
\begin{equation}
\boldsymbol{P}^{(k)}
=
[
\boldsymbol{p}_1^{(k)},
\boldsymbol{p}_2^{(k)},
\ldots,
\boldsymbol{p}_T^{(k)}
],
\quad k=1,2,\ldots,K.
\label{MC_sampling}
\end{equation}
By repeating this sampling process $K$ times, we obtain a set of stochastic prompt instances that captures distributional variability in the learned prompt space.

\textbf{Normality-aware anomaly prompt modeling.}
Because anomaly detection identifies statistical deviations from normal conditions, we propose a normality-aware prompt design that models anomaly distributions as parametric deviations from normal distributions serving as reference distributions. For each $i$th token in a prompt sequence of length $T$, the normal prompt token $\boldsymbol{\varphi}_i$ and anomalous prompt token $\boldsymbol{\psi}_i$ are modeled as diagonal multivariate Gaussian distributions:
\begin{equation}
\begin{aligned}
\boldsymbol{\varphi}_i
&\sim
\mathcal{N}
(
\boldsymbol{\mu}_i^{n,\theta},
\boldsymbol{\Sigma}_i^{n,\theta}
),\\
\boldsymbol{\psi}_i
&\sim
\mathcal{N}
(
\boldsymbol{\mu}_i^{a,\theta},
\boldsymbol{\Sigma}_i^{a,\theta}
),
\quad i=1,2,\ldots,T.
\end{aligned}
\end{equation}
The anomalous distribution parameters are modeled as learnable deviations from the corresponding normal distribution:
\begin{equation}
\begin{aligned}
\boldsymbol{\mu}_i^{a,\theta}
&=
\boldsymbol{\mu}_i^{n,\theta}
+
\boldsymbol{\delta}_{\mu,i}^{\theta},\\
\boldsymbol{\Sigma}_i^{a,\theta}
&=
\boldsymbol{\Sigma}_i^{n,\theta}
+
\boldsymbol{\Delta}_{\Sigma,i}^{\theta}.
\label{normality-aware}
\end{aligned}
\end{equation}
where $\boldsymbol{\delta}_{\mu,i}^{\theta} \in \mathbb{R}^d$ and $\boldsymbol{\Delta}_{\Sigma,i}^{\theta}$ denote the learnable mean and diagonal covariance deviations, respectively. Through this design, anomalous prompts encode relational abnormality by learning statistical deviations from their corresponding normal references.

To effectively incorporate this distribution-based modeling into prompts, we design a prompt structure that integrates normal and anomaly distributions by leveraging how the text encoder aggregates semantic information at the final \texttt{<eos>} token. We adopt an object-agnostic approach for ZSAD, and the proposed distributional prompts are defined as follows:
\begin{equation}
\begin{aligned}
\boldsymbol{\mathit{\Phi}}
&=
[\boldsymbol{\varphi}_1,\boldsymbol{\varphi}_2,\ldots,\boldsymbol{\varphi}_T]
\oplus
[\boldsymbol{e}_{obj}],\\
\boldsymbol{\mathit{\Psi}}
&=
[\boldsymbol{\varphi}_1,\boldsymbol{\varphi}_2,\ldots,\boldsymbol{\varphi}_T]
\oplus
[\boldsymbol{\psi}_1,\boldsymbol{\psi}_2,\ldots,\boldsymbol{\psi}_T]
\oplus
[\boldsymbol{e}_{dmg},\boldsymbol{e}_{obj}],
\label{prompt_combination}
\end{aligned}
\end{equation}
where $\oplus$ denotes the concatenation operator, and $\boldsymbol{e}_{dmg}$ and $\boldsymbol{e}_{obj}$ are the pre-trained word embeddings for ``\texttt{damaged}'' and ``\texttt{object}'', respectively. For each prompt state, $K$ samples are generated according to Eq.~\ref{MC_sampling}.

\subsection{Hierarchical visual design}

\textbf{Hierarchical distributional prompt.}
Visual anomalies occur at various spatial scales, ranging from semantic anomalies at the entire image level to local defects in specific regions. To learn specialized representations for the global and local levels of this visual hierarchy, we extend the distribution-based prompts to two levels. First, the distributional token set at the global level is defined as
\begin{equation}
\boldsymbol{\mathcal{D}}_g
:=
\left\{
(\boldsymbol{\varphi}_{g,i},
\boldsymbol{\psi}_{g,i})
\right\}_{i=1}^{T},
\end{equation}
where each $\boldsymbol{\varphi}_{g,i}$ and $\boldsymbol{\psi}_{g,i}$ follows a diagonal Gaussian distribution as defined above. The normal prompt parameters are denoted by $\boldsymbol{\mu}_{g,i}^{n,\theta}$ and $\boldsymbol{\Sigma}_{g,i}^{n,\theta}$, while the anomalous parameters $\boldsymbol{\mu}_{g,i}^{a,\theta}$ and $\boldsymbol{\Sigma}_{g,i}^{a,\theta}$ are defined as deviations from the corresponding normal distribution according to Eq.~\ref{normality-aware}. The global distribution is aligned with image-level representations and specialized for classification. From these distribution parameters, $K$ prompts are generated using Eq.~\ref{prompt_combination} to form the global prompt sample space:
\begin{equation}
\boldsymbol{\mathcal{S}}_g
:=
\left\{
(
\boldsymbol{\mathit{\Phi}}_g^{(k)},
\boldsymbol{\mathit{\Psi}}_g^{(k)}
)
\right\}_{k=1}^{K}.
\end{equation}

Similarly, the distributional token set at the local level is defined as
\begin{equation}
\boldsymbol{\mathcal{D}}_l
:=
\left\{
(
\boldsymbol{\varphi}_{l,i},
\boldsymbol{\psi}_{l,i}
)
\right\}_{i=1}^{T},
\end{equation}
where each $\boldsymbol{\varphi}_{l,i}$ and $\boldsymbol{\psi}_{l,i}$ follows a diagonal Gaussian distribution. The anomalous distribution is again defined as a parametric deviation from its corresponding normal distribution based on Eq.~\ref{normality-aware}. The local distribution is specialized for pixel-level anomaly localization.

The $K$ local prompt samples generated from this distribution using Eq.~\ref{prompt_combination} form the following sample space:
\begin{equation}
\boldsymbol{\mathcal{S}}_l
:=
\left\{
(
\boldsymbol{\mathit{\Phi}}_l^{(k)},
\boldsymbol{\mathit{\Psi}}_l^{(k)}
)
\right\}_{k=1}^{K}.
\end{equation}

The prototype text embeddings corresponding to each level are constructed as follows. First, text embeddings are extracted from the prompt sample spaces $\boldsymbol{\mathcal{S}}_g$ and $\boldsymbol{\mathcal{S}}_l$ using the text encoder $\mathcal{E}_t$:
\begin{equation}
\begin{aligned}
\boldsymbol{F}_g^{n},\boldsymbol{F}_g^{a}
&=
\mathcal{E}_{t}(\boldsymbol{\mathcal{S}}_g),\\
\boldsymbol{F}_l^{n},\boldsymbol{F}_l^{a}
&=
\mathcal{E}_{t}(\boldsymbol{\mathcal{S}}_l).
\end{aligned}
\end{equation}
Each output is represented as a matrix composed of $K$ text embeddings:
\begin{equation}
\begin{aligned}
\boldsymbol{F}_g^{n}
&=
[
\boldsymbol{f}_g^{n,(1)},
\boldsymbol{f}_g^{n,(2)},
\ldots,
\boldsymbol{f}_g^{n,(K)}
]^T
\in\mathbb{R}^{K\times d},\\
\boldsymbol{F}_g^{a}
&=
[
\boldsymbol{f}_g^{a,(1)},
\boldsymbol{f}_g^{a,(2)},
\ldots,
\boldsymbol{f}_g^{a,(K)}
]^T
\in\mathbb{R}^{K\times d},\\
\boldsymbol{F}_l^{n}
&=
[
\boldsymbol{f}_l^{n,(1)},
\boldsymbol{f}_l^{n,(2)},
\ldots,
\boldsymbol{f}_l^{n,(K)}
]^T
\in\mathbb{R}^{K\times d},\\
\boldsymbol{F}_l^{a}
&=
[
\boldsymbol{f}_l^{a,(1)},
\boldsymbol{f}_l^{a,(2)},
\ldots,
\boldsymbol{f}_l^{a,(K)}
]^T
\in\mathbb{R}^{K\times d}.
\end{aligned}
\end{equation}
The prototype text embeddings are computed by averaging the vectors in each matrix:
\begin{equation}
\begin{aligned}
\boldsymbol{f}_g^{n}
&=
\frac{1}{K}
\sum_{k=1}^{K}
\boldsymbol{f}_g^{n,(k)},
\quad
\boldsymbol{f}_g^{a}
=
\frac{1}{K}
\sum_{k=1}^{K}
\boldsymbol{f}_g^{a,(k)},\\
\boldsymbol{f}_l^{n}
&=
\frac{1}{K}
\sum_{k=1}^{K}
\boldsymbol{f}_l^{n,(k)},
\quad
\boldsymbol{f}_l^{a}
=
\frac{1}{K}
\sum_{k=1}^{K}
\boldsymbol{f}_l^{a,(k)}.
\end{aligned}
\end{equation}
This prototype generation aggregates the stochastic variations introduced by Monte Carlo sampling into a compact semantic representation for each state. Because the learned prompt distributions are image-agnostic, the $K$ prompt samples can be encoded and averaged once after training to construct the global and local normal/anomalous prototypes. These prototype text embeddings can then be cached and reused for all query images during inference, without repeating Monte Carlo sampling and text encoding for every test image.

The embeddings at each level are denoted as
\begin{equation}
\begin{aligned}
\boldsymbol{f}_g
&\coloneqq
\{
\boldsymbol{f}_g^{n},
\boldsymbol{f}_g^{a}
\},
\quad
\boldsymbol{f}_l
\coloneqq
\{
\boldsymbol{f}_l^{n},
\boldsymbol{f}_l^{a}
\},\\
\boldsymbol{F}_g
&\coloneqq
\{
\boldsymbol{F}_g^{n},
\boldsymbol{F}_g^{a}
\},
\quad
\boldsymbol{F}_l
\coloneqq
\{
\boldsymbol{F}_l^{n},
\boldsymbol{F}_l^{a}
\}.
\end{aligned}
\end{equation}
The resulting global and local prototypes provide level-specific textual representations for image-level anomaly detection and pixel-level localization, respectively.

\textbf{Hierarchical visual feature adaptation.}
To effectively utilize the different levels of visual knowledge contained in each layer of $\mathcal{E}_v$, we use multi-scale patch features extracted from intermediate layers. From the selected layer set $\mathcal{L}=\{l_1,l_2,l_3,l_4\}$, we extract patch-level features $\boldsymbol{H}_{l_i}\in\mathbb{R}^{N\times d}$ from each layer $l_i$, and adapt each feature using a transformation function $\mathcal{A}_{l_i}^{\theta}(\cdot)$:
\begin{equation}
\begin{aligned}
\boldsymbol{H}_{l_i}'
&=
\mathcal{A}_{l_i}^{\theta}
(\boldsymbol{H}_{l_i}),\\
\boldsymbol{H}_{l_i}'
&\in
\mathbb{R}^{N\times d}.
\end{aligned}
\end{equation}
where $\mathcal{A}_{l_i}^{\theta}$ is a layer-specific local adapter. This adaptive transformation enables the features from each layer to more effectively capture anomaly patterns at the corresponding level. The anomaly map is computed using the adapted patch features according to Eqs.~\ref{anomaly_map} and \ref{agg_anomaly_map}.

For the global representation, the final projection output $\boldsymbol{v}_{pro}$ from $\mathcal{E}_v$ is adapted using a global adapter $\mathcal{A}_g^\theta(\cdot)$:
\begin{equation}
\boldsymbol{v}_g
=
\mathcal{A}_g^\theta
(\boldsymbol{v}_{pro}),
\quad
\boldsymbol{v}_{pro}
\in
\mathbb{R}^{d}.
\label{global adapter}
\end{equation}
This yields an adapted global feature $\boldsymbol{v}_g$, which serves as a semantic representation specialized for anomaly detection.

\subsection{Knowledge-guided contrastive learning}
\label{section: knowledge-guided}

\begin{figure}[pos=htbp]
    \centering
    \adjustbox{clip=False}{
        \includegraphics[width=0.9\textwidth]{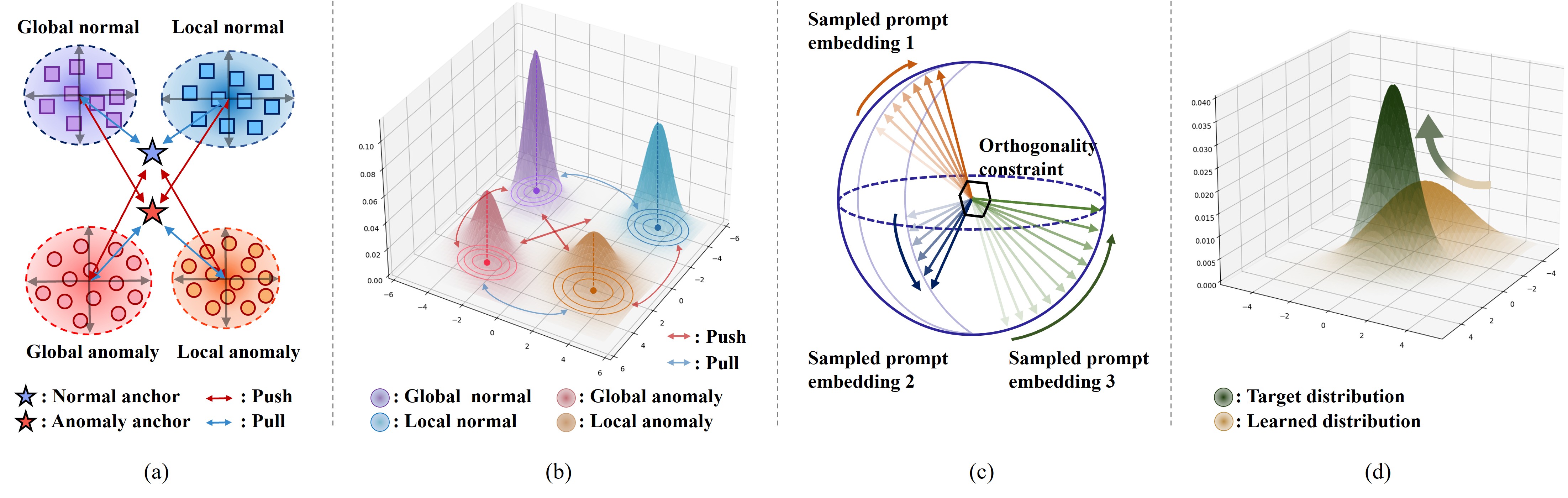}
    }
    \vspace{-1mm}
    \caption{Illustration of the comprehensive learning strategy proposed in HiDist. (a) Knowledge-guided contrastive learning. (b) Distributional loss. (c) Orthogonality constraint. (d) Kullback--Leibler (KL) divergence regularization.}
    \label{fig: comprehensive learning}
\end{figure}

\textbf{Knowledge anchor.}
Although hierarchical distribution-based prompts enable specialized learning at each level, independently optimizing the global and local branches may weaken semantic consistency across levels. To address this issue, we employ a contrastive learning strategy that uses the adapted global visual feature $\boldsymbol{v}_g$ as a shared knowledge anchor, as illustrated in Fig.~\ref{fig: comprehensive learning}(a). This encourages the global and local distributional prompts to remain semantically aligned with the global visual representation while preserving their level-specific roles.

% Figure 3

\textbf{Visual hierarchy alignment.}
Based on the knowledge anchor, we introduce knowledge-guided contrastive learning for aligning prototype text embeddings and learning explicit boundaries between normal and anomalous states. Positive and negative prompts are distinguished according to the image labels. For normal images ($y=0$):
\begin{equation}
\begin{aligned}
\mathcal{L}_{triplet}^{global}(y=0)
&=
\max
(
0,
m_g
-
sim(\boldsymbol{v}_g,\boldsymbol{f}_g^n)
+
sim(\boldsymbol{v}_g,\boldsymbol{f}_g^a)
),\\
\mathcal{L}_{triplet}^{local}(y=0)
&=
\max
(
0,
m_l
-
sim(\boldsymbol{v}_g,\boldsymbol{f}_l^n)
+
sim(\boldsymbol{v}_g,\boldsymbol{f}_l^a)
).
\end{aligned}
\end{equation}
For anomalous images ($y=1$):
\begin{equation}
\begin{aligned}
\mathcal{L}_{triplet}^{global}(y=1)
&=
\max
(
0,
m_g
-
sim(\boldsymbol{v}_g,\boldsymbol{f}_g^a)
+
sim(\boldsymbol{v}_g,\boldsymbol{f}_g^n)
),\\
\mathcal{L}_{triplet}^{local}(y=1)
&=
\max
(
0,
m_l
-
sim(\boldsymbol{v}_g,\boldsymbol{f}_l^a)
+
sim(\boldsymbol{v}_g,\boldsymbol{f}_l^n)
).
\end{aligned}
\end{equation}
Here, $m_g$ and $m_l$ are the margin hyperparameters for the global and local branches, respectively. The visual hierarchy loss is computed as
\begin{equation}
\mathcal{L}_{hi}
=
\mathcal{L}_{triplet}^{global}(y)
+
\mathcal{L}_{triplet}^{local}(y),
\end{equation}
where $y$ is the ground-truth label of the corresponding image.

\subsection{Comprehensive learning strategy}

\begin{figure}[pos=htbp]
    \centering
    \adjustbox{clip=False}{
        \includegraphics[width=0.9\textwidth]{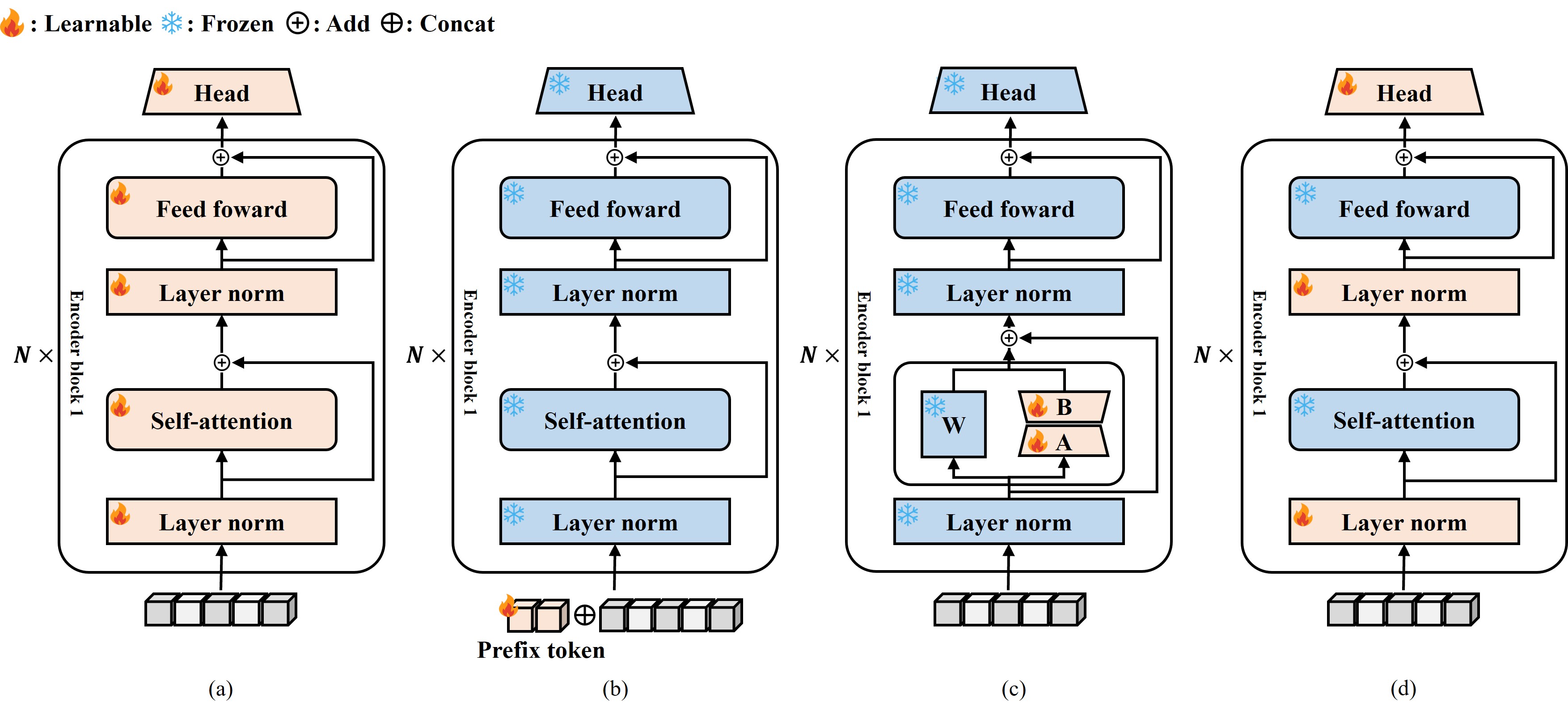}
    }
    \vspace{-1mm}
    \caption{Comparison of text encoder tuning strategies. (a) Full tuning. (b) Prefix tuning~\citep{li2021prefix}. (c) Low-Rank Adaptation~\citep{hu2022lora}. (d) LayerNorm and head tuning.}
    \label{fig: text encoder tuning}
\end{figure}

To effectively learn hierarchical distribution-based prompts, we employ a comprehensive learning strategy that combines task-specific objectives with four auxiliary loss terms and selective fine-tuning of the text encoder, as presented in Figs.~\ref{fig: comprehensive learning} and \ref{fig: text encoder tuning}.

\textbf{Regularization of hierarchical distributions.}
As shown in Fig.~\ref{fig: comprehensive learning}(b), to promote cross-level consistency between distributions representing the same semantic state while maintaining separation between normal and anomalous distributions at each hierarchical level, we define a distributional loss using the squared 2-Wasserstein distance~\citep{gabriel2019computational}. Specifically, the global and local normal distributions are encouraged to remain close to each other, as are the global and local anomalous distributions, while normal and anomalous distributions within the same level are encouraged to remain separated:
\begin{equation}
\begin{aligned}
\mathcal{L}_{dist}
&=
\mathrm{softplus}
(
m
-
W_2^2(\boldsymbol{\mathcal{D}}_g^n,\boldsymbol{\mathcal{D}}_g^a)
+
W_2^2(\boldsymbol{\mathcal{D}}_g^n,\boldsymbol{\mathcal{D}}_l^n)
)\\
&\quad+
\mathrm{softplus}
(
m
-
W_2^2(\boldsymbol{\mathcal{D}}_l^n,\boldsymbol{\mathcal{D}}_l^a)
+
W_2^2(\boldsymbol{\mathcal{D}}_g^a,\boldsymbol{\mathcal{D}}_l^a)
).
\end{aligned}
\end{equation}
Here, $W_2^2(\cdot,\cdot)$ denotes the average token-wise squared 2-Wasserstein distance between the corresponding diagonal Gaussian distributions in two distributional token sets. $\boldsymbol{\mathcal{D}}_g^n$, $\boldsymbol{\mathcal{D}}_g^a$, $\boldsymbol{\mathcal{D}}_l^n$, and $\boldsymbol{\mathcal{D}}_l^a$ denote the normal and anomalous distributional token sets at the global and local levels, respectively, and $m$ is the shared margin hyperparameter. The softplus function provides a smooth ranking objective that encourages same-state distributions across the two hierarchical levels to remain close while increasing the separation between normal and anomalous distributions within each level.

\textbf{Orthogonality constraints for prompt diversity.}
\label{orthogonal loss}
As shown in Fig.~\ref{fig: comprehensive learning}(c), we apply orthogonality constraints to encourage sufficient coverage of the textual embedding space among the $K$ Monte Carlo prompt samples. The orthogonal loss is defined as
\begin{equation}
\begin{aligned}
\mathcal{L}_{orth}
&=
\|
\boldsymbol{F}_g^{n,{norm}}
(\boldsymbol{F}_g^{n,{norm}})^T
-
\boldsymbol{I}_K
\|_F^2\\
&\quad+
\|
\boldsymbol{F}_g^{a,{norm}}
(\boldsymbol{F}_g^{a,{norm}})^T
-
\boldsymbol{I}_K
\|_F^2\\
&\quad+
\|
\boldsymbol{F}_l^{n,{norm}}
(\boldsymbol{F}_l^{n,{norm}})^T
-
\boldsymbol{I}_K
\|_F^2\\
&\quad+
\|
\boldsymbol{F}_l^{a,{norm}}
(\boldsymbol{F}_l^{a,{norm}})^T
-
\boldsymbol{I}_K
\|_F^2.
\end{aligned}
\end{equation}
where $\boldsymbol{F}_g^{n,{norm}}$, $\boldsymbol{F}_g^{a,{norm}}$, $\boldsymbol{F}_l^{n,{norm}}$, and $\boldsymbol{F}_l^{a,{norm}}$ are the row-wise $l_2$-normalized feature matrices, and $\boldsymbol{I}_K$ is the $K\times K$ identity matrix.

\textbf{Prior regularization of prompt distributions.}
As shown in Fig.~\ref{fig: comprehensive learning}(d), we regularize the learned Gaussian distributions using Kullback--Leibler (KL) divergence~\citep{kullback1997information}:
\begin{equation}
\begin{aligned}
\mathcal{L}_{kl}
&=
\frac{1}{4}
\Big[
KL(
\boldsymbol{\mathcal{D}}_g^n
|
\mathcal{N}(\boldsymbol{0},\boldsymbol{I})
)
+
KL(
\boldsymbol{\mathcal{D}}_g^a
|
\mathcal{N}(\boldsymbol{0},\boldsymbol{I})
)\\
&\quad+
KL(
\boldsymbol{\mathcal{D}}_l^n
|
\mathcal{N}(\boldsymbol{0},\boldsymbol{I})
)
+
KL(
\boldsymbol{\mathcal{D}}_l^a
|
\mathcal{N}(\boldsymbol{0},\boldsymbol{I})
)
\Big].
\end{aligned}
\end{equation}
Here, each KL term is obtained by averaging the KL divergences of the token-wise diagonal Gaussian distributions in the corresponding distributional set. This regularization discourages degenerate distribution parameters by constraining the learned token distributions relative to a standard normal prior, thereby promoting stable distributional learning.

\textbf{Text encoder fine-tuning.}
To enhance distributional learning while preserving the pre-trained knowledge of the text encoder, we apply selective fine-tuning to only the layer-normalization parameters and the final projection head:
\begin{equation}
\theta_{text}
:=
\left\{
\gamma_\ell,\beta_\ell
\right\}_{\ell=1}^{L_t}
\cup
\left\{
\boldsymbol{W}_{head},
\boldsymbol{b}_{head}
\right\},
\end{equation}
where $L_t$ denotes the number of text-encoder layers, $\gamma_\ell$ and $\beta_\ell$ denote the layer-normalization parameters of each layer, and $\boldsymbol{W}_{head}$ and $\boldsymbol{b}_{head}$ represent the weight and bias of the final projection head, respectively. As illustrated in Fig.~\ref{fig: text encoder tuning}(d), only the parameter set $\theta_{text}$ is updated during training.

% Figure 4

\textbf{Task-specific optimization.}
We incorporate task-specific loss functions for both classification and localization. Image-level anomaly classification is optimized using binary cross-entropy loss:
\begin{equation}
\mathcal{L}_{cls}
=
\mathcal{L}_{bce}
(
A_{cls}
(
\boldsymbol{v}_g,
\boldsymbol{f}_g^a
),
y
),
\end{equation}
where $y$ is the ground-truth label and $A_{cls}$ is the predicted anomaly score computed as in Eq.~\ref{anomaly_cls}. For pixel-level anomaly localization, the segmentation loss combines focal loss~\citep{lin2017focal} and Dice loss~\citep{milletari2016v}:
\begin{equation}
\begin{aligned}
\mathcal{L}_{seg}
=
\sum_{l_i\in\mathcal{L}}
\big(
&
\mathcal{L}_{focal}
(
\boldsymbol{A}_{seg}^{(i)}
(
\boldsymbol{H}_{l_i}',
\boldsymbol{f}_l
),
\boldsymbol{M}
)\\
+
&
\mathcal{L}_{dice}
(
\boldsymbol{A}_{seg}^{(i)}
(
\boldsymbol{H}_{l_i}',
\boldsymbol{f}_l
),
\boldsymbol{M}
)
\big).
\end{aligned}
\end{equation}
Here, $\boldsymbol{M}$ denotes the ground-truth anomaly mask of the auxiliary training image.

The total loss is defined as
\begin{equation}
\mathcal{L}_{total}
=
\mathcal{L}_{cls}
+
\mathcal{L}_{seg}
+
\mathcal{L}_{hi}
+
\mathcal{L}_{dist}
+
\lambda_1\mathcal{L}_{orth}
+
\lambda_2\mathcal{L}_{kl},
\end{equation}
where $\lambda_1$ and $\lambda_2$ are regularization hyperparameters.

\textbf{Training and inference.}
During training, the prompt distribution parameters, hierarchical visual adapters, and selectively tunable text-encoder parameters are jointly optimized using $\mathcal{L}_{total}$. During inference, the learned image-agnostic prompt distributions are sampled once to construct the global and local normal/anomalous text prototypes, which are then cached. For each query image, only the visual features and their hierarchical adaptations are computed, and image-level anomaly detection and pixel-level localization are performed using the cached global and local prototypes according to Eqs.~\ref{anomaly_cls}--\ref{agg_anomaly_map}.

\section{Experiments}
\subsection{Experimental setup}
\textbf{Datasets.}
Our evaluation spans both industrial and medical domains. In the industrial domain, we employ MVTec-AD~\citep{bergmann2019mvtec}, VisA~\citep{zou2022spot}, and DTD-Synthetic~\citep{aota2023zero}. MVTec-AD consists of 15 categories of high-resolution industrial images and covers diverse structural and textural anomaly types. VisA contains 12 categories of real-world objects with fine-grained and subtle defects. DTD-Synthetic consists of 12 categories with diverse synthetically generated texture anomalies, providing a complementary setting for evaluating generalization to unseen defect patterns.

In the medical domain, we utilize CVC-ColonDB~\citep{tajbakhsh2015automated}, CVC-ClinicDB~\citep{bernal2015wm}, Kvasir~\citep{jha2019kvasir}, and Endo~\citep{hicks2021endotect}. These datasets consist of endoscopic images collected across different clinical settings and provide pixel-level lesion annotations, enabling cross-dataset evaluation of anomaly localization under variations in anatomical appearance and lesion morphology.

For evaluation in the industrial domain, we assess cross-dataset generalization by training on the test set of MVTec-AD and evaluating on VisA and DTD-Synthetic, and vice versa by training on the VisA test set and evaluating on MVTec-AD. In the medical domain, we train on CVC-ClinicDB and evaluate on CVC-ColonDB, Kvasir, and Endo, and vice versa by training on CVC-ColonDB and evaluating on CVC-ClinicDB.

\textbf{Evaluation Metrics.}
To comprehensively assess anomaly detection performance, we adopted separate metrics for image-level classification and pixel-level segmentation. For image-level evaluation, we report  the area under the receiver operating 
characteristic curve (AUROC), average precision (AP), and the maximum F1 score (F1-max) at the optimal threshold. For the pixel-level evaluation, we employed AUROC, per-region overlap (PRO)~\citep{bergmann2020uninformed}, and F1-max to capture both detection accuracy and spatial alignment.

\textbf{Implementation details.}
We adopted the SigLIP 2 model (\texttt{so400m-patch16-512}) as the backbone. The prompt length $T$ was set to 13 for both normal and anomalous classes at the global and local levels. To effectively capture hierarchical visual representations, we extract multi-scale patch tokens from the middle layers $\mathcal{L}={5,12,19,26}$ of the vision encoder. Local adapters were initialized with the same structure and weights as SigLIP 2, whereas the global adapter followed the structure of SigLIP 2's projection head. The sampling size $K$ for each distribution was set to 20. The Gaussian filter parameter $\sigma$ was set to 8, and the regularization coefficients were $\lambda_{orth} = 0.001$ and $\lambda_{kl} = 0.0001$. We optimized the model using Adam with a learning rate of 0.0001 and a weight decay of $1 \times 10^{-5}$. Training was performed for 15 epochs on a single NVIDIA RTX 4090 GPU.

\subsection{Experimental results}
For comprehensive evaluation, we included baseline results from OpenCLIP (\texttt{ViT-L-14-336}) and SigLIP 2 (\texttt{so400m-patch16-512}). We further compared HiDist with representative ZSAD methods, including WinCLIP~\citep{jeong2023winclip}, APRIL-GAN~\citep{chen2023april}, AnomalyCLIP~\citep{zhou2024anomalyclip}, AdaCLIP~\citep{cao2024adaclip}, and Bayes-PFL~\citep{qu2025bayesian}.

\textbf{Comparison on industrial domain.}
Tables~\ref{table1} and~\ref{table2} report the image- and pixel-level ZSAD results on the industrial datasets. At the image level, HiDist achieves the best performance across all datasets and evaluation metrics. Compared with Bayes-PFL, HiDist improves AUROC by 4.2\%p, 3.9\%p, and 2.1\%p on MVTec-AD, VisA, and DTD-Synthetic, respectively, while also achieving higher AP and F1-max scores.

% Table 1 - Performance comparison of ZSAS approaches
\begin{table}[pos=htbp]
\centering
\caption{Comparison of image-level ZSAD results with state-of-the-art methods on industrial datasets. Bolded values indicate the best performance, and underlined values indicate the second-best.}
\label{table1}
% \resizebox{0.97\textwidth}{!}
{
\begin{tabular}{@{}lccccccccc@{}}
\toprule
\multirow{2.5}{*}{\textbf{Method}}
& \multicolumn{3}{c}{MVTec-AD}
& \multicolumn{3}{c}{VisA}
& \multicolumn{3}{c}{DTD-Synthetic} \\
\cmidrule(lr){2-4} \cmidrule(lr){5-7} \cmidrule(lr){8-10}
& AUROC & AP & F1-max
& AUROC & AP & F1-max
& AUROC & AP & F1-max \\
\midrule
OpenCLIP
& 63.7 & 84.1 & 85.0
& 59.4 & 68.8 & 74.7
& 55.6 & 81.3 & 85.7 \\

SigLIP 2
& 60.5 & 81.9 & 84.6
& 59.7 & 68.7 & 74.3
& 83.1 & 93.3 & 90.1 \\

WinCLIP
& 90.4 & 95.6 & \underline{92.7}
& 75.6 & 78.8 & 78.2
& 95.1 & 97.7 & 94.1 \\

APRIL-GAN
& 90.8 & 95.4 & 91.7
& 83.0 & 84.6 & 81.6
& 85.5 & 94.0 & 89.0 \\

AnomalyCLIP
& \underline{91.5} & \underline{96.2} & \underline{92.7}
& 81.4 & 84.9 & 80.7
& 93.7 & 97.3 & 93.9 \\

AdaCLIP
& 91.2 & 95.9 & 92.3
& 81.7 & 84.0 & 81.1
& \underline{96.8} & 98.3 & \underline{96.1} \\

Bayes-PFL
& 90.5 & 95.8 & 92.5
& \underline{85.4} & \underline{88.5} & \underline{82.7}
& 95.7 & \underline{98.5} & 95.8 \\

\midrule
\textbf{HiDist (Ours)}
& \textbf{94.7} & \textbf{97.4} & \textbf{93.6}
& \textbf{89.3} & \textbf{91.1} & \textbf{85.9}
& \textbf{97.8} & \textbf{99.2} & \textbf{96.8} \\
\bottomrule
\end{tabular}}
\end{table}

% Table 2
\begin{table}[pos=htbp]
\centering
\caption{Comparison of pixel-level ZSAD results with state-of-the-art methods on industrial datasets. Bolded values indicate the best performance, and underlined values indicate the second-best.}
\label{table2}
\renewcommand{\arraystretch}{1.2}
% \resizebox{0.97\textwidth}{!}
{
\begin{tabular}{@{}lccccccccc@{}}
\toprule
\multirow{2.5}{*}{\textbf{Method}}
& \multicolumn{3}{c}{MVTec-AD}
& \multicolumn{3}{c}{VisA}
& \multicolumn{3}{c}{DTD-Synthetic} \\
\cmidrule(lr){2-4} \cmidrule(lr){5-7} \cmidrule(lr){8-10}
& AUROC & PRO & F1-max
& AUROC & PRO & F1-max
& AUROC & PRO & F1-max \\
\midrule
OpenCLIP
& 66.3 & 27.0 & 9.8
& 69.3 & 35.2 & 3.2
& 74.9 & 37.9 & 6.0 \\

SigLIP 2
& 66.7 & 27.2 & 10.0
& 69.0 & 34.9 & 3.2
& 75.7 & 39.2 & 6.3 \\

WinCLIP
& 82.3 & 61.9 & 24.8
& 73.2 & 51.1 & 9.0
& 79.5 & 51.5 & 16.1 \\

APRIL-GAN
& 87.6 & 44.0 & 43.3
& 94.2 & 86.6 & 32.3
& 96.5 & 41.5 & 65.4 \\

AnomalyCLIP
& 91.0 & 81.9 & 39.0
& 95.3 & 85.1 & 27.3
& 97.8 & 91.1 & 61.5 \\

AdaCLIP
& 82.5 & 15.0 & 35.9
& 95.5 & 77.8 & \underline{35.2}
& \underline{98.9} & 79.8 & \underline{74.2} \\

Bayes-PFL
& \underline{91.6} & \textbf{87.4} & \underline{47.6}
& \textbf{96.1} & \textbf{90.0} & \textbf{35.3}
& 98.3 & \underline{95.0} & 68.1 \\

\midrule
\textbf{HiDist (Ours)}
& \textbf{92.2} & \underline{87.0} & \textbf{49.0}
& \underline{95.7} & \underline{88.3} & 31.1
& \textbf{99.2} & \textbf{95.6} & \textbf{75.2} \\
\bottomrule
\end{tabular}}
\end{table}

%Table3
\begin{table}[pos=htbp]
\centering
\caption{Comparison of ZSAD results with state-of-the-art methods on medical datasets. Bolded values indicate the best performance, and underlined values indicate the second-best.}
\label{table3}
\resizebox{\textwidth}{!}{
\begin{tabular}{@{}lcccccccccccc@{}}
\toprule
\multirow{2.5}{*}{\textbf{Method}}
& \multicolumn{3}{c}{CVC-ColonDB}
& \multicolumn{3}{c}{CVC-ClinicDB}
& \multicolumn{3}{c}{Kvasir}
& \multicolumn{3}{c}{Endo} \\
\cmidrule(lr){2-4} \cmidrule(lr){5-7} \cmidrule(lr){8-10} \cmidrule(lr){11-13}
& AUROC & PRO & F1-max
& AUROC & PRO & F1-max
& AUROC & PRO & F1-max
& AUROC & PRO & F1-max \\
\midrule
OpenCLIP
& 63.6 & 30.1 & 18.7
& 67.0 & 31.4 & 24.1
& 67.1 & 26.6 & 34.3
& 70.3 & 36.6 & 35.2 \\

SigLIP 2
& 64.1 & 29.5 & 18.9
& 67.2 & 32.4 & 24.3
& 67.3 & 26.1 & 34.6
& 70.5 & 36.4 & 35.4 \\

WinCLIP
& 64.8 & 28.4 & 21.0
& 70.3 & 32.5 & 27.2
& 69.7 & 24.5 & 35.9
& 68.2 & 28.3 & 32.9 \\

APRIL-GAN
& 78.4 & 28.0 & 32.5
& 85.4 & 41.0 & 44.2
& 80.5 & 17.3 & 48.4
& 84.1 & 32.3 & 51.0 \\

AnomalyCLIP
& 88.5 & 79.3 & 52.4
& 93.0 & 81.6 & 64.2
& 93.2 & 59.9 & 70.7
& 94.1 & 86.9 & 72.6 \\

AdaCLIP
& 88.0 & 63.9 & \underline{61.3}
& 94.4 & 73.5 & \underline{75.5}
& 94.6 & 26.2 & 76.2
& 95.2 & 72.8 & 77.2 \\

Bayes-PFL
& \underline{90.3} & \underline{85.6} & 60.4
& \underline{95.5} & \underline{88.9} & 72.9
& \underline{95.6} & \textbf{80.3} & \underline{77.4}
& \underline{96.2} & \underline{92.1} & \underline{78.4} \\

\midrule
\textbf{HiDist (Ours)}
& \textbf{93.5} & \textbf{90.2} & \textbf{69.3}
& \textbf{96.1} & \textbf{91.0} & \textbf{80.1}
& \textbf{97.5} & \underline{78.5} & \textbf{84.4}
& \textbf{97.7} & \textbf{95.7} & \textbf{84.2} \\
\bottomrule
\end{tabular}}
\end{table}

\begin{figure}[pos=htbp]
    \centering
    \includegraphics[width=0.97\linewidth]{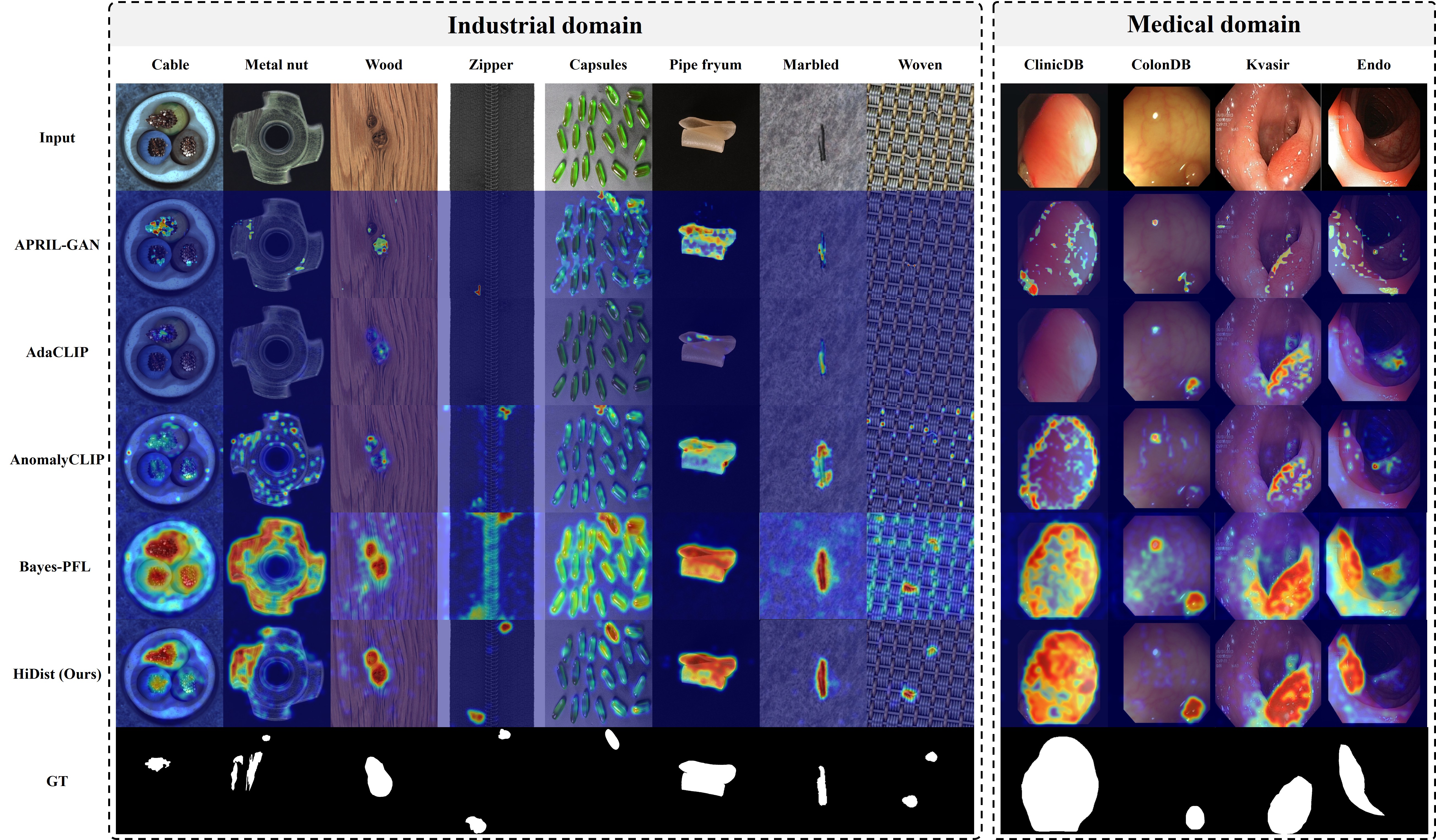}
    \caption{Qualitative comparison of anomaly localization results across various ZSAD methods. The left panel shows results on industrial datasets, where the first four columns correspond to MVTec-AD~\citep{bergmann2019mvtec}, followed by two columns each for VisA~\citep{zou2022spot} and DTD-Synthetic~\citep{aota2023zero}. The right panel presents results on medical datasets in the order of CVC-ClinicDB~\citep{bernal2015wm}, CVC-ColonDB~\citep{tajbakhsh2015automated}, Kvasir~\citep{jha2019kvasir}, and Endo~\citep{hicks2021endotect}.}
    \label{fig:main comparison}
\end{figure}

At the pixel level, the results are more competitive. HiDist achieves the best AUROC and F1-max on MVTec-AD and the best results across all three metrics on DTD-Synthetic, whereas Bayes-PFL obtains the highest PRO on MVTec-AD and outperforms HiDist across all three metrics on VisA. In particular, Bayes-PFL exceeds HiDist on VisA by 0.4\%p, 1.7\%p, and 4.2\%p in AUROC, PRO, and F1-max, respectively. Although Bayes-PFL achieves higher pixel-level scores on VisA, the qualitative examples in the left panel of Fig.~\ref{fig:main comparison} show that HiDist often produces more spatially concentrated anomaly responses, indicating that threshold-based localization metrics and visual compactness capture complementary aspects of anomaly localization. Overall, HiDist maintains strong pixel-level performance across the industrial datasets while showing a more compact localization tendency in several qualitative cases.

\textbf{Comparison on medical domain.}
Table~\ref{table3} reports the pixel-level ZSAD results on the medical datasets. HiDist achieves the best performance in 11 of the 12 reported metrics across CVC-ColonDB, CVC-ClinicDB, Kvasir, and Endo, demonstrating consistently strong localization performance across different datasets. Notably, HiDist achieves the highest AUROC and F1-max on all four datasets and the highest PRO on three of the four datasets, indicating that its performance gains are consistent across complementary localization metrics rather than being concentrated in a single evaluation criterion. The improvements are particularly pronounced in F1-max, where HiDist exceeds the strongest competing results by 8.0\%p, 4.6\%p, 7.0\%p, and 5.8\%p on CVC-ColonDB, CVC-ClinicDB, Kvasir, and Endo, respectively. The qualitative results in the right panel of Fig.~\ref{fig:main comparison} further show that HiDist consistently identifies lesion regions across images with diverse appearances and lesion scales. Overall, HiDist demonstrates strong cross-dataset generalization for medical anomaly localization.

\subsection{Ablation studies}
We conducted ablation studies to validate the effectiveness of each core component of HiDist. Experimental results are reported using all evaluation metrics, with scores averaged across datasets in the industrial (MVTec-AD, VisA, DTD-Synthetic) and medical domains (CVC-ClinicDB, CVC-ColonDB, Kvasir, Endo).

\textbf{Effectiveness of normality-aware prompt modeling and visual hierarchy.}
Table~\ref{tab:ablation1} evaluates the contributions of normality-aware prompt modeling and visual hierarchy. In the industrial domain, applying either component individually improves performance over the baseline, while combining both achieves the best overall results. Specifically, image-level AUROC increases from 90.2 to 93.9, and pixel-level PRO increases from 87.1 to 90.3. A similar trend is observed in the medical domain, where PRO improves from 82.5 to 88.9 when both components are incorporated. These results indicate that the two components provide complementary benefits for anomaly detection and localization.

\begin{figure}[pos=htbp]
    \centering
    \includegraphics[width=0.7\linewidth]{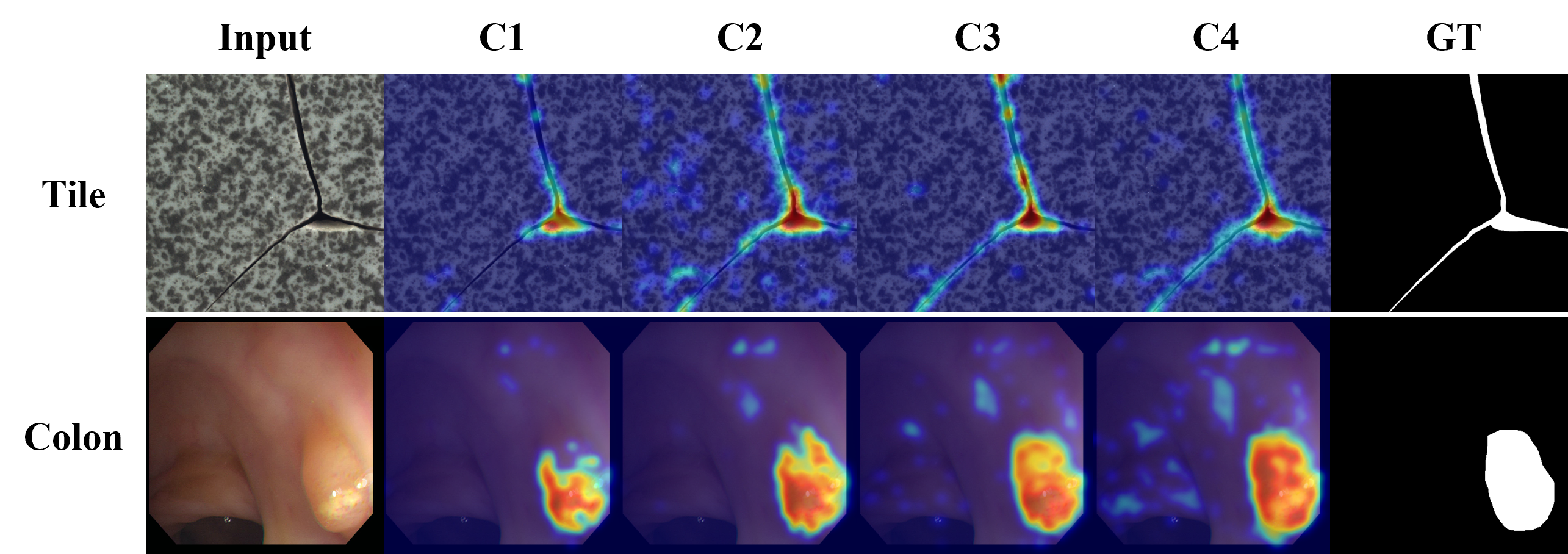}
    \caption{Visual comparison of anomaly localization performance under different configurations of normality-aware prompting and visual hierarchy. Each row shows examples from MVTec-AD~\citep{bergmann2019mvtec} and CVC-ClinicDB~\citep{bernal2015wm}. Columns represent four experimental conditions: (C1) baseline (neither method), (C2) normality-aware prompting only, (C3) visual hierarchy only, and (C4) both methods applied.}
    \label{fig:condition_compare}
\end{figure}

Fig.~\ref{fig:condition_compare} compares anomaly localization results under different design configurations. The baseline setting (C1) fails to detect subtle cracks in the tile class and misidentifies anomalous regions that closely resemble normal areas in the colon class. Applying only normality-aware prompting (C2) or only visual hierarchy (C3) improves localization performance, enabling more accurate detection of subtle cracks and anomalous regions. When both components are integrated (C4), the model more completely captures the anomalous regions, demonstrating a clear synergistic effect between the two design elements.

\begin{table}[pos=htbp]
\centering
\caption{Ablation of normality-aware prompt modeling and visual hierarchy across industrial and medical domains. Results are shown for image-level (classification) and pixel-level (segmentation) tasks, under different combinations of the two proposed components. Bolded values indicate the best performance in each setting.}
\label{tab:ablation1}
% \resizebox{0.67\textwidth}{!}
{
\begin{tabular}{@{}lllcccc@{}}
\toprule
\textbf{Domain} & \textbf{Level} & \textbf{Normality-aware} & \textbf{Visual hierarchy}  & \textbf{AUROC} & \textbf{AP/PRO} & \textbf{F1-max} \\
\midrule
\multirow{9}{*}{Industrial} & \multirow{4}{*}{Image}
& \multicolumn{1}{c}{\ding{55}} & \multicolumn{1}{c}{\ding{55}} 
& 90.2 & 94.3 & 90.7 \\
& & \multicolumn{1}{c}{\ding{51}} & \multicolumn{1}{c}{\ding{55}} 
& 92.9 & 95.3 & 91.4 \\
& & \multicolumn{1}{c}{\ding{55}} & \multicolumn{1}{c}{\ding{51}} 
&  93.4 & 95.5 & 92.1\\
& & \multicolumn{1}{c}{\ding{51}} & \multicolumn{1}{c}{\ding{51}} 
& \textbf{93.9} & \textbf{95.9} & \textbf{92.1} \\
\cmidrule{2-7}

& \multirow{4}{*}{Pixel}
& \multicolumn{1}{c}{\ding{55}} & \multicolumn{1}{c}{\ding{55}}
& 95.3 & 87.1 & 49.8 \\
& & \multicolumn{1}{c}{\ding{51}} & \multicolumn{1}{c}{\ding{55}} 
& 95.5 & 87.7 & 50.9 \\
& & \multicolumn{1}{c}{\ding{55}} & \multicolumn{1}{c}{\ding{51}}
& 95.6 & 88.9 & 51.3\\
& & \multicolumn{1}{c}{\ding{51}} & \multicolumn{1}{c}{\ding{51}}
& \textbf{95.7} & \textbf{90.3} & \textbf{51.8} \\
\midrule

\multirow{4}{*}{Medical} & \multirow{4}{*}{Pixel}
& \multicolumn{1}{c}{\ding{55}} & \multicolumn{1}{c}{\ding{55}}
& 94.5 & 82.5 & 77.5 \\
& & \multicolumn{1}{c}{\ding{51}} & \multicolumn{1}{c}{\ding{55}}
& 95.4 & 84.2 & 78.0 \\
& & \multicolumn{1}{c}{\ding{55}} & \multicolumn{1}{c}{\ding{51}}
& 95.8 & 86.8 & 79.0\\
& & \multicolumn{1}{c}{\ding{51}} & \multicolumn{1}{c}{\ding{51}}
& \textbf{96.2} & \textbf{88.9} & \textbf{79.5} \\
\bottomrule
\end{tabular}}
\end{table}

\textbf{Effectiveness of distribution-based prompt modeling.}
Table~\ref{tab:ablation2} evaluates the effects of distribution-based prompt modeling using Monte Carlo sampling. For comparison, we implemented an ensemble approach with the same prompt structure, where $K=20$ prompts were initialized as independent learnable vectors rather than sampled from distributions. In the industrial domain, distribution-based modeling yielded a 2.0\%p improvement in image-level AUROC over the learnable prompt ensemble, with similar gains observed in pixel-level metrics. This result suggests that learning prompt distributions can provide more effective stochastic variation than independently optimizing a fixed set of prompt vectors. A similar trend was observed in the medical domain, where distribution-based prompts achieved a 1.8\%p improvement in PRO. Overall, these results support the effectiveness of distributional prompt modeling compared with an equally sized ensemble of independently learnable prompts.

\begin{table}[pos=htbp]
\centering
\caption{Ablation of distribution-based prompt on industrial and medical domains. Bolded values indicate the best performance.}
\label{tab:ablation2}
% \resizebox{0.67\textwidth}{!}
{
\begin{tabular}{@{}lllcccc@{}}
\toprule
\textbf{Domain} & \textbf{Level}& \textbf{Distribution-aware}  & \textbf{AUROC} & \textbf{AP/PRO} & \textbf{F1-max} \\
\midrule
\multirow{5}{*}{Industrial} & \multirow{2}{*}{Image}
& \multicolumn{1}{c}{\ding{55}} 
& 91.9 & 93.9 & 91.6 \\
& & \multicolumn{1}{c}{\ding{51}} 
& \textbf{93.9} & \textbf{95.9} & \textbf{92.1} \\
\cmidrule(lr){2-6}
& \multirow{2}{*}{Pixel}
& \multicolumn{1}{c}{\ding{55}}
& 95.5 & 89.2 & 51.6 \\
& & \multicolumn{1}{c}{\ding{51}}
& \textbf{95.7} & \textbf{90.3} & \textbf{51.8} \\
\midrule
\multirow{2}{*}{Medical} & \multirow{2}{*}{Pixel}
& \multicolumn{1}{c}{\ding{55}}
& 95.5 & 87.1 & 78.2 \\
& & \multicolumn{1}{c}{\ding{51}}
& \textbf{96.2} & \textbf{88.9} & \textbf{79.5} \\
\bottomrule
\end{tabular}}
\end{table}

\begin{table}[pos=htbp]
\centering
\caption{Ablation of comprehensive learning strategy on industrial and medical domains. Bolded values indicate the best performance.}
\label{tab:ablation3}
% \resizebox{0.6\textwidth}{!}
{
\begin{tabular}{@{}lllccc@{}}
\toprule
\textbf{Domain} & \textbf{Level} & \textbf{w/o setting}  & \textbf{AUROC} & \textbf{AP/PRO} & \textbf{F1-max} \\
\midrule
\multirow{11}{*}{Industrial} & \multirow{5}{*}{Image}
& w/o visual hierarchy loss & 93.1 & 95.2 & 91.6 \\
& & w/o distributional loss & 92.6 & 94.8 & 91.8 \\
& & w/o orthogonal loss & 91.9 & 95.0 & 91.6 \\
& & w/o KL divergence loss & 93.3 & 95.5 & 92.0 \\
& & \textbf{Ours} & \textbf{93.9} & \textbf{95.9} & \textbf{92.1}\\
\cmidrule(lr){2-6}
& \multirow{5}{*}{Pixel}
& w/o visual hierarchy loss & 95.1 & 86.6 & 51.1 \\
& & w/o distributional loss & 95.4 & 87.3 & 51.1 \\
& & w/o orthogonal loss & 95.4 & 88.1 & 50.6 \\
& & w/o KL divergence loss & 95.5 & 88.4 & 51.1 \\
& & \textbf{Ours} & \textbf{95.7} & \textbf{90.3} & \textbf{51.8} \\
\midrule

\multirow{5}{*}{Medical} & \multirow{5}{*}{Pixel}
& w/o visual hierarchy loss & 95.5 & 84.9 & 78.5 \\
& & w/o distributional loss & 96.0 & 85.9 & 78.1 \\
& & w/o orthogonal loss & 95.5 & 84.2 & 77.7 \\
& & w/o KL divergence loss & 95.6 & 85.7 & 78.3 \\
& & \textbf{Ours} & \textbf{96.2} & \textbf{88.9} & \textbf{79.5} \\
\bottomrule
\end{tabular}}
\end{table}

\textbf{Effectiveness of comprehensive learning strategy.}
To assess the contribution of the proposed learning objectives, we individually removed the four auxiliary loss components: visual hierarchy, distributional, orthogonality, and KL divergence losses. As shown in Table~\ref{tab:ablation3}, removing any of these objectives degrades the overall performance, although their relative contributions differ across evaluation levels. In the industrial domain, the orthogonality loss has the largest effect on image-level AUROC, with its removal reducing the score by 2.0\%p, followed by the distributional loss with a 1.3\%p decrease. For pixel-level localization, removing the visual hierarchy and distributional losses decreases PRO by 3.7\%p and 3.0\%p, respectively. This result suggests that the objectives associated with hierarchical alignment and distributional organization play a particularly important role in spatial anomaly localization, whereas sample diversity induced by the orthogonality constraint contributes more strongly to image-level discrimination.

A similar pattern is observed in the medical domain, where the complete objective achieves the best overall performance. Removing the orthogonality loss produces the largest PRO degradation of 4.7\%p, followed by the visual hierarchy loss with a 4.0\%p decrease. Excluding the distributional and KL divergence losses also reduces PRO by 3.0\%p and 3.2\%p, respectively. The consistent degradation observed when each objective is removed indicates that the four auxiliary losses contribute complementary regularization to the proposed framework rather than relying on a single dominant objective.

\begin{figure}[pos=htbp]
    \centering
    \includegraphics[width=0.6\linewidth]{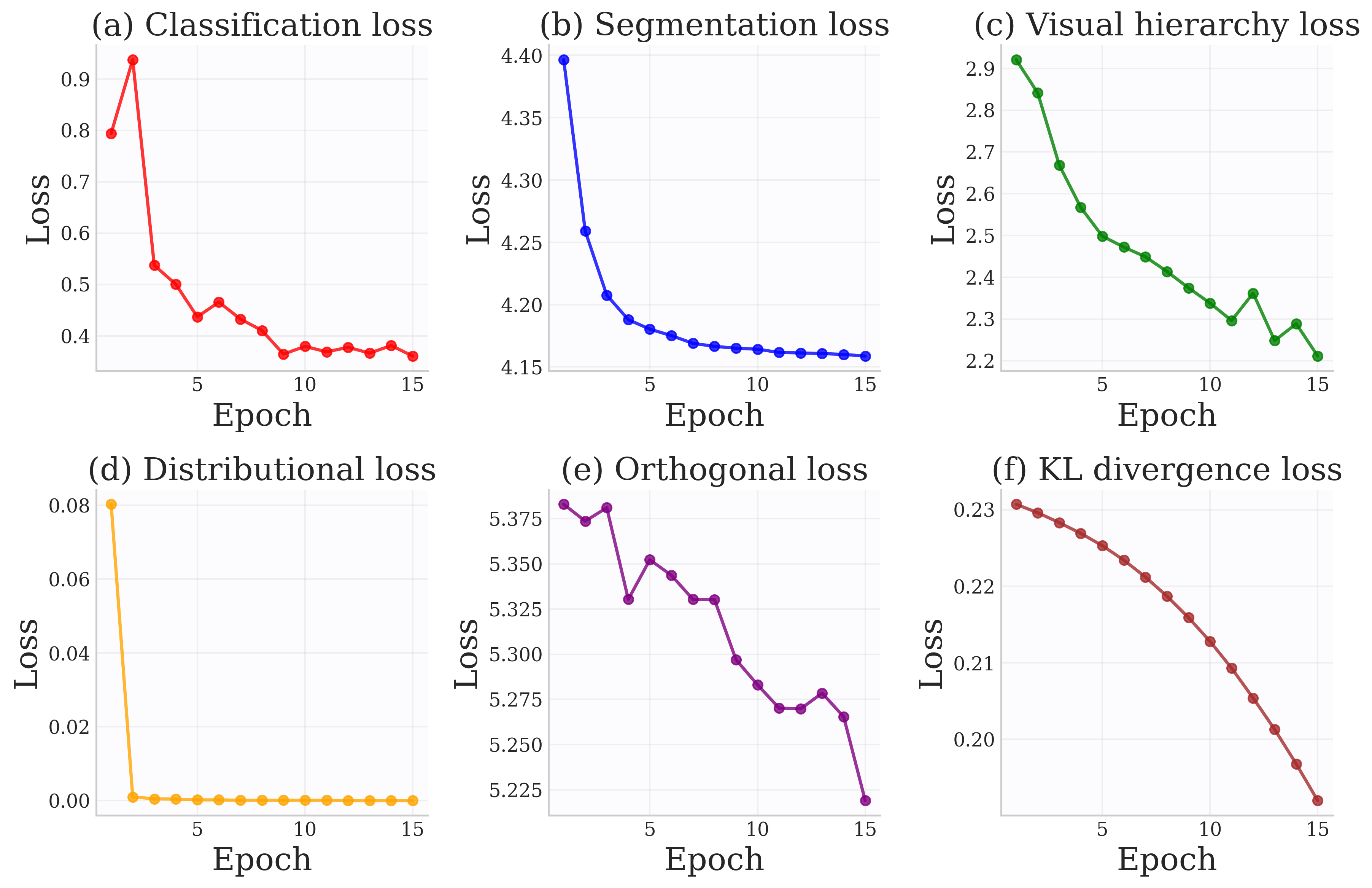}
    \caption{Training convergence curves for each loss component. The plots show: (a) classification loss, (b) segmentation loss, (c) visual hierarchy loss, (d) distributional loss, (e) orthogonality loss, and (f) KL divergence loss.}
    \label{fig:loss graph}
\end{figure}

Fig.~\ref{fig:loss graph} further illustrates the optimization behavior of the individual loss components. The classification and segmentation losses decrease and stabilize as training progresses, while the visual hierarchy, orthogonality, and KL divergence losses show overall decreasing trends with different convergence rates. In particular, the distributional loss rapidly decreases to a very small value during the early stage of training and remains low thereafter, suggesting that its relative-distance constraint is largely satisfied early in optimization. Together with the ablation results, these curves indicate that the auxiliary objectives are jointly optimized without exhibiting unstable divergence during training.

\textbf{Effectiveness of visual adapters.}
To evaluate the proposed global and local adapters, we removed each component individually. As shown in Table~\ref{tab:ablation4}, both adapters contribute to performance, but their relative importance differs across domains. In the industrial domain, removing the local adapter causes substantial degradation in pixel-level localization, reducing PRO and F1-max by 11.1\%p and 27.6\%p, respectively, while removing the global adapter decreases PRO by 10.0\%p. In contrast, the global adapter is more influential in the medical domain, where its removal reduces AUROC, PRO, and F1-max by 9.6\%p, 22.3\%p, and 27.9\%p, respectively. These results suggest complementary roles of the two adapters, with local adaptation contributing more strongly to fine-grained industrial localization and global adaptation being particularly important for medical anomaly localization.

\begin{table}[pos=htbp]
\centering
\caption{Ablation results for global and local visual adapters on industrial and medical domains. Bolded values indicate the best performance.}
\label{tab:ablation4}
% \resizebox{0.8\linewidth}{!}
{
\begin{tabular}{@{}ccccccc@{}}
\toprule
\textbf{Domain} & \textbf{Level} & \textbf{Global adapter} & \textbf{Local adapters} & \textbf{AUROC} & \textbf{AP/PRO} & \textbf{F1-max} \\
\midrule
\multirow{7}{*}{Industrial} & \multirow{3}{*}{Image}
& \multicolumn{1}{c}{\ding{51}} & \multicolumn{1}{c}{\ding{55}} & 90.5 & 93.2 & 91.4 \\
& & \multicolumn{1}{c}{\ding{55}} & \multicolumn{1}{c}{\ding{51}} & 93.2 & 95.6 & 91.7 \\
& & \multicolumn{1}{c}{\ding{51}} & \multicolumn{1}{c}{\ding{51}} & \textbf{93.9} & \textbf{95.9} & \textbf{92.1} \\
\cmidrule(lr){2-7}
& \multirow{3}{*}{Pixel}
& \multicolumn{1}{c}{\ding{51}} & \multicolumn{1}{c}{\ding{55}} & 91.1 & 79.2 & 24.2 \\
& & \multicolumn{1}{c}{\ding{55}} & \multicolumn{1}{c}{\ding{51}} & 94.3 & 80.3 & 48.6  \\
& & \multicolumn{1}{c}{\ding{51}} & \multicolumn{1}{c}{\ding{51}} & \textbf{95.7} & \textbf{90.3} & \textbf{51.8} \\
\midrule
\multirow{3}{*}{Medical} & \multirow{3}{*}{Pixel}
& \multicolumn{1}{c}{\ding{51}} & \multicolumn{1}{c}{\ding{55}} & 95.2 & 83.2 & 77.6\\
& & \multicolumn{1}{c}{\ding{55}} & \multicolumn{1}{c}{\ding{51}} & 86.6 & 66.6 & 51.6 \\
& & \multicolumn{1}{c}{\ding{51}} & \multicolumn{1}{c}{\ding{51}} & \textbf{96.2} & \textbf{88.9} & \textbf{79.5}\\
\bottomrule
\end{tabular}}
\end{table}

\begin{figure}[pos=htbp]
    \centering
    \includegraphics[width=0.8\linewidth]{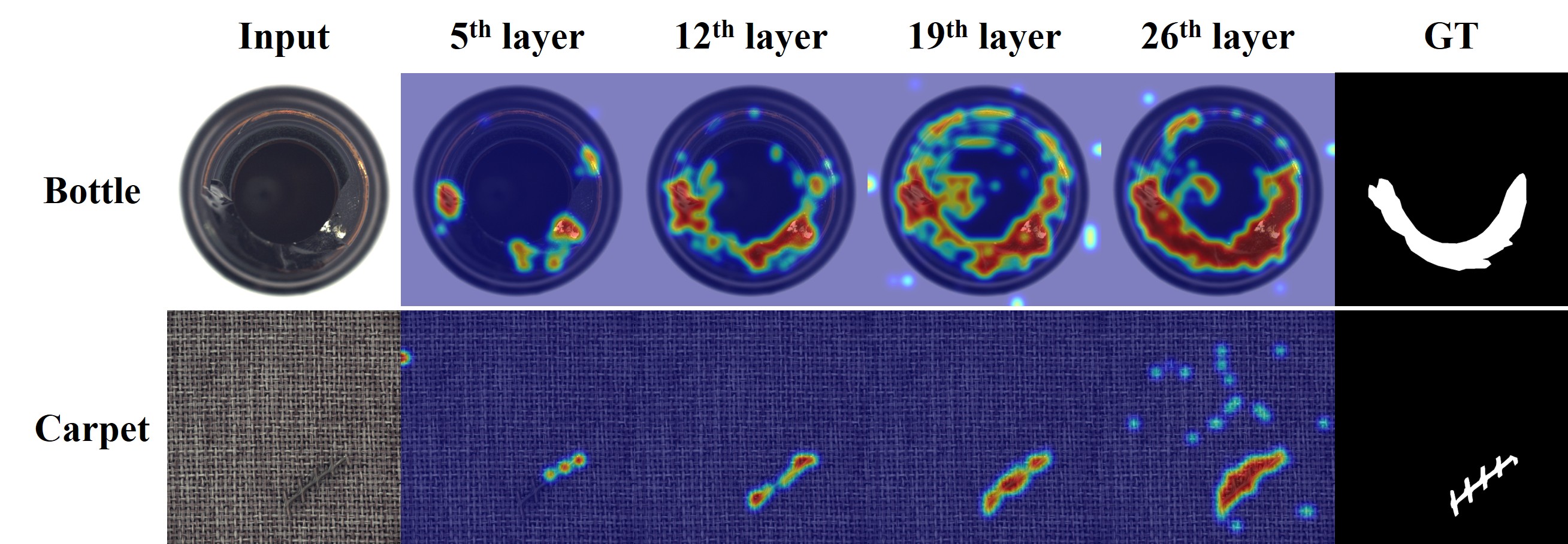}
    \caption{Anomaly localization results using multi-scale patch features extracted from the 5th, 12th, 19th, and 26th layers of the vision encoder. Each column shows the anomaly map corresponding to one of the selected layers for an input from MVTec-AD~\citep{bergmann2019mvtec}, highlighting the effect of hierarchical features.}
    \label{fig:middle layer}
\end{figure}

\textbf{Effectiveness of multi-scale patch features.}
We evaluated how performance changes as patch features from additional intermediate layers are incorporated. As shown in Table~\ref{tab:ablation5}, performance consistently improves with the number of feature levels in both industrial and medical domains, with particularly pronounced gains in pixel-level localization. Fig.~\ref{fig:middle layer} provides further insight into this trend: features from earlier layers capture fine-grained local details with relatively concentrated responses, whereas features from later layers provide broader contextual information but may introduce additional background responses. These observations support the complementary use of multi-scale patch features for anomaly localization.

\begin{table}[pos=htbp]
\centering
\caption{Ablation of multi-scale patch features on industrial and medical domains. Bolded values indicate the best performance.}
\label{tab:ablation5}
% \resizebox{0.7\linewidth}{!}
{
\begin{tabular}{@{}ccccccc@{}}
\toprule
\textbf{Domain} & \textbf{Level} & \textbf{Number of patch features} & \textbf{AUROC} & \textbf{AP/PRO} & \textbf{F1-max} \\
\midrule
\multirow{7}{*}{Industrial} & \multirow{3}{*}{Image}
& 1 & 90.9 & 93.0 & 91.6 \\
& & 2 & 93.6 & 95.7 & 91.8 \\
& & 4 & \textbf{93.9} & \textbf{95.9} & \textbf{92.1} \\
\cmidrule(lr){2-6}
& \multirow{3}{*}{Pixel}
& 1 & 93.9 & 79.3 & 43.2 \\
& & 2 & 95.0 & 83.8 & 48.2 \\
& & 4 & \textbf{95.7} & \textbf{90.3} & \textbf{51.8} \\
\midrule
\multirow{3}{*}{Medical} & \multirow{3}{*}{Pixel}
& 1 & 94.3 & 77.7 & 76.1 \\
& & 2 & 95.6 & 82.1 & 77.4 \\
& & 4 & \textbf{96.2} & \textbf{88.9} & \textbf{79.5} \\
\bottomrule
\end{tabular}}
\end{table}

\textbf{Effectiveness of text encoder fine-tuning.}
To assess the impact of different fine-tuning strategies for the text encoder, we compared five approaches: (1) freezing all parameters, (2) full fine-tuning of the encoder, (3) prefix tuning~\citep{li2021prefix}, (4) low-rank adaptation (LoRA)~\citep{hu2022lora}, and (5) selective tuning of the LayerNorm parameters and the final projection head. As shown in Table~\ref{tab:ablation6}, in the industrial domain, full fine-tuning, prefix tuning, and LoRA all result in lower image-level performance than the frozen baseline, whereas selective tuning of the LayerNorm parameters and the final projection head achieves the best results across all image-level and pixel-level metrics. This indicates that more extensive adaptation of the text encoder does not necessarily benefit the proposed distributional prompt learning framework. In the medical domain, the adaptation strategies generally improve some pixel-level metrics over the frozen baseline, with selective LayerNorm and head tuning again achieving the best overall performance. These results support selective text-encoder adaptation as an effective strategy for incorporating learned distributional prompts while retaining the pre-trained text representation.

\begin{table}[pos=htbp]
\centering
\caption{Ablation of text encoder fine-tuning strategies on industrial and medical domains. Bolded values indicate the best performance.}
\label{tab:ablation6}
% \resizebox{0.7\textwidth}{!}
{
\begin{tabular}{@{}lllccc@{}}
\toprule
\textbf{Domain} & \textbf{Level} & \textbf{Fine-tuning method} & \textbf{AUROC} & \textbf{AP/PRO} & \textbf{F1-max} \\
\midrule
\multirow{11}{*}{Industrial} & \multirow{5}{*}{Image}
& Frozen & 93.6 & 95.7 & 92.0 \\
& & Full tuning & 92.7 & 94.6 & 91.8 \\
& & Prefix-tuning & 67.8 & 79.4 & 85.9 \\
& & LoRA & 92.0 & 94.7 & 91.5 \\
& & LayerNorm and head tuning & \textbf{93.9} & \textbf{95.9} & \textbf{92.1} \\
\cmidrule(lr){2-6}
& \multirow{5}{*}{Pixel}
& Frozen & 95.3 & 87.2 & 50.6 \\
& & Full tuning & 95.3 & 88.7 & 51.4 \\
& & Prefix-tuning & 95.4 & 89.1 & 51.3\\
& & LoRA & 95.6 & 88.9 & 51.4\\
& & LayerNorm and head tuning & \textbf{95.7} & \textbf{90.3} & \textbf{51.8} \\
\midrule
\multirow{5}{*}{Medical} & \multirow{5}{*}{Pixel}
& Frozen & 95.4 & 85.4 & 78.4 \\
& & Full tuning & 95.7 & 87.9 & 78.3 \\
& & Prefix-tuning & 95.9 & 87.2 & 79.0\\
& & LoRA & 95.9 & 88.2 & 79.4\\
& & LayerNorm and head tuning & \textbf{96.2} & \textbf{88.9} & \textbf{79.5} \\
\bottomrule
\end{tabular}}
\end{table}

\section{Discussion}
\subsection{Analysis of hierarchical prompt parameters}
\begin{figure}[pos=htbp]
    \centering
    \includegraphics[width=0.8\linewidth]{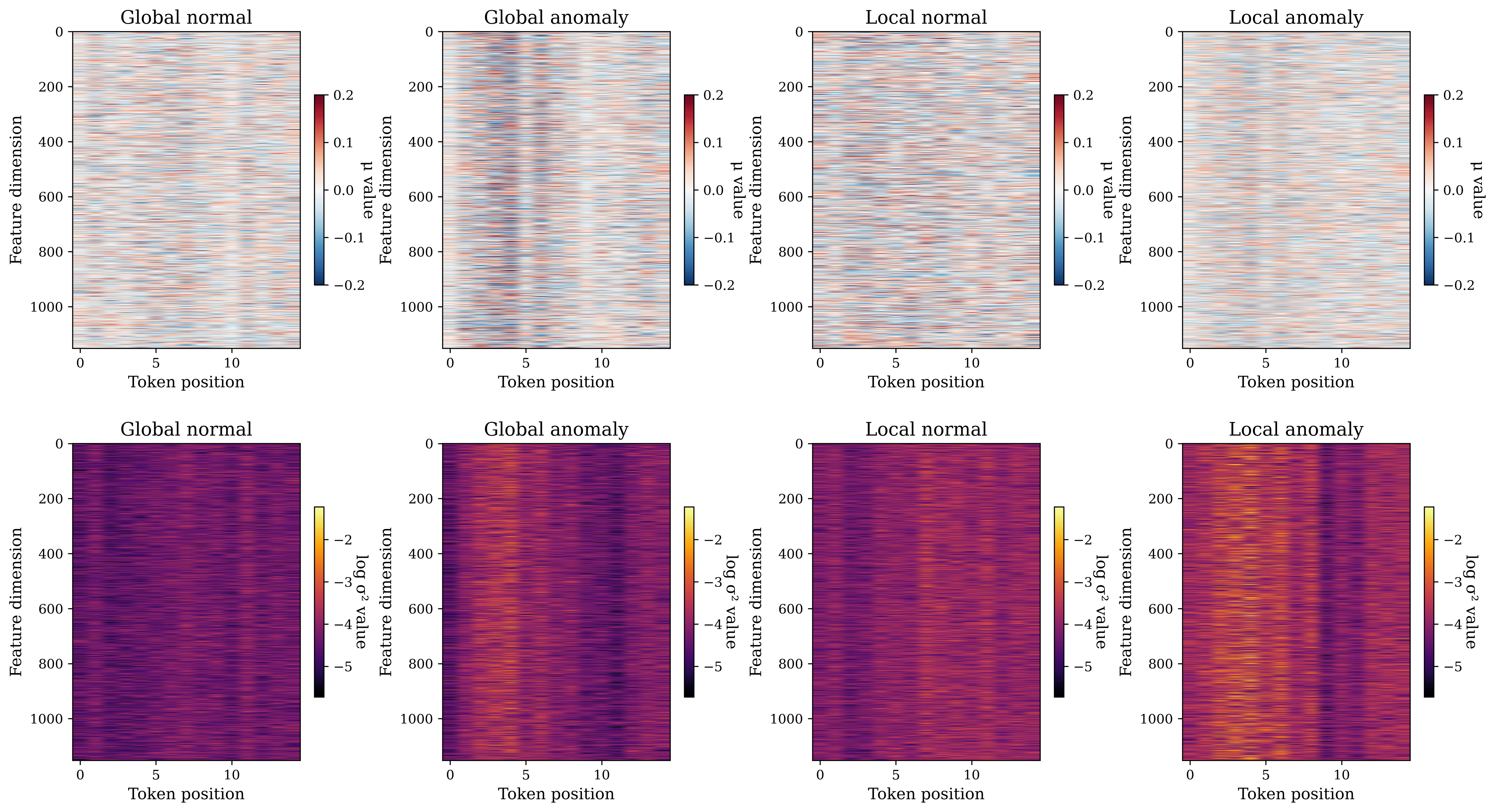}
    \caption{Visualization of learned prompt distribution parameters. The top row shows the mean vectors ($\boldsymbol{\mu}$) and the bottom row shows the corresponding log-variance values ($\log \boldsymbol{\sigma}^2$) for each prompt token.}
    \label{fig:distribution para}
\end{figure}

Fig.~\ref{fig:distribution para} visualizes the Gaussian prompt distributions learned on VisA~\citep{zou2022spot}, showing the mean vectors $\boldsymbol{\mu}$ and log-variance values $\log \boldsymbol{\sigma}^2$ across prompt-token positions. The four prompt types---global normal, global anomaly, local normal, and local anomaly---exhibit distinct parameter patterns, indicating that the global/local and normal/anomalous branches learn differentiated prompt-space representations.

The normal prompt distributions generally exhibit lower log-variance values, whereas the anomaly prompt distributions show greater variability across token positions, with the local anomaly prompts displaying particularly pronounced variations. This suggests that the learned anomaly prompts retain greater stochastic flexibility than their normal counterparts, especially at the local level. Overall, the parameter patterns provide qualitative evidence that the proposed distributional prompt formulation learns specialized global/local and normal/anomalous representations rather than collapsing to similar distributions.

\subsection{Analysis of distributional differences across hierarchical prompts}
\begin{figure}[pos=htbp]
    \centering
    \begin{subfigure}{0.32\textwidth}
        \centering
        \includegraphics[width=\textwidth]{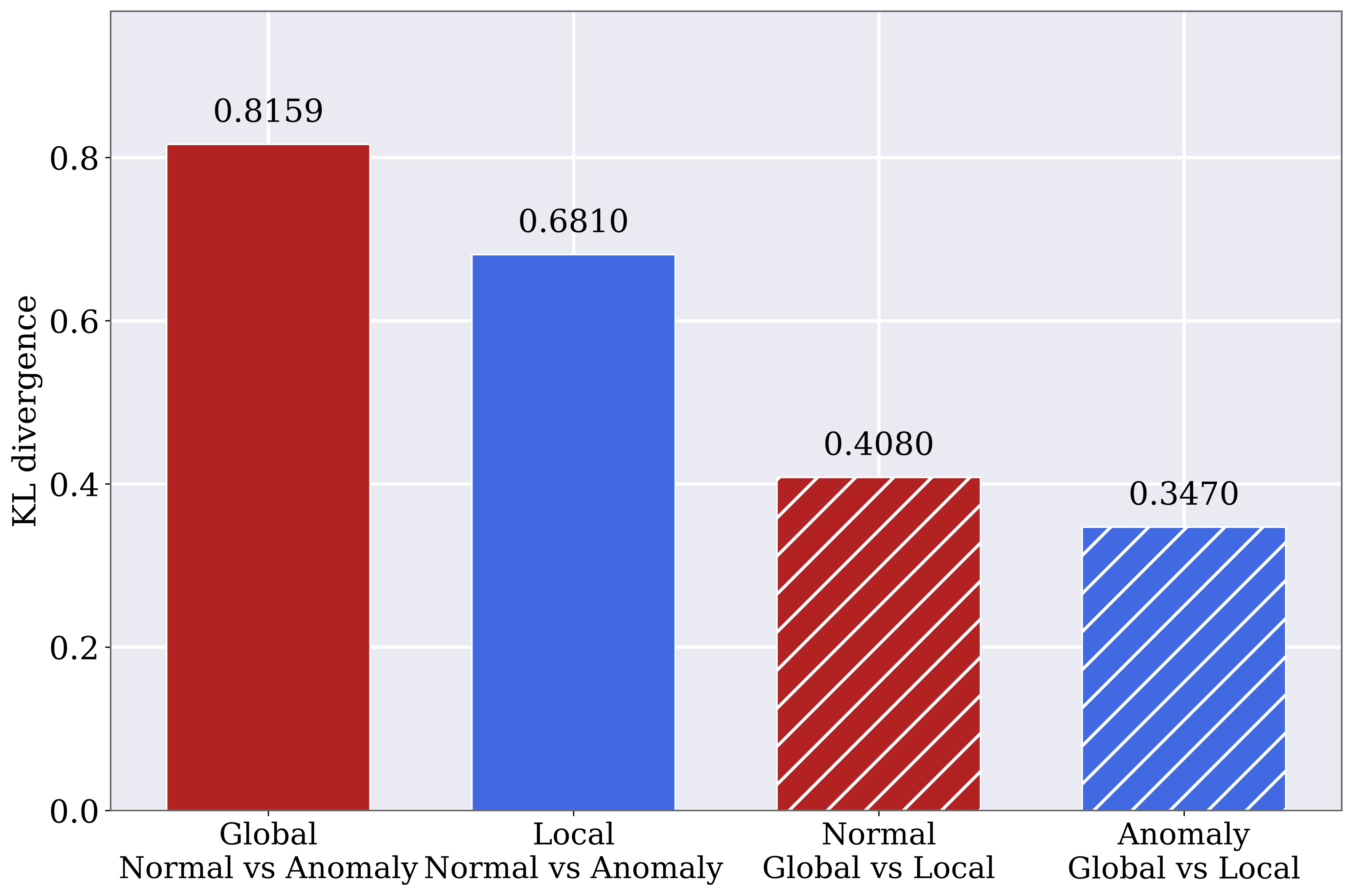}
        \caption{KL divergence}
    \end{subfigure}
    % \hfill
    \begin{subfigure}{0.32\textwidth}
        \centering
        \includegraphics[width=\textwidth]{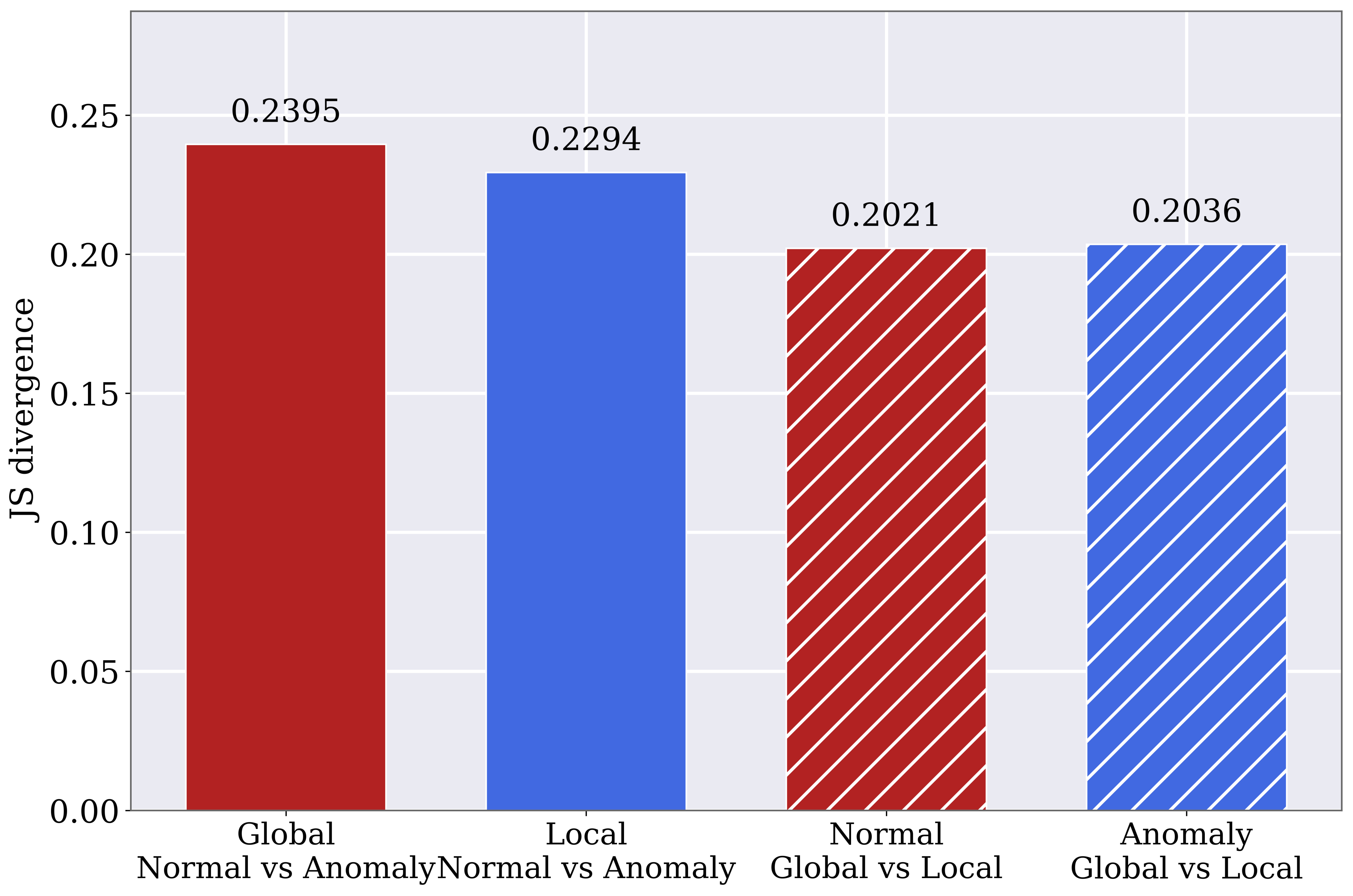}
        \caption{Wasserstein distance}
    \end{subfigure}
    \begin{subfigure}{0.32\textwidth}
        \centering
        \includegraphics[width=\textwidth]{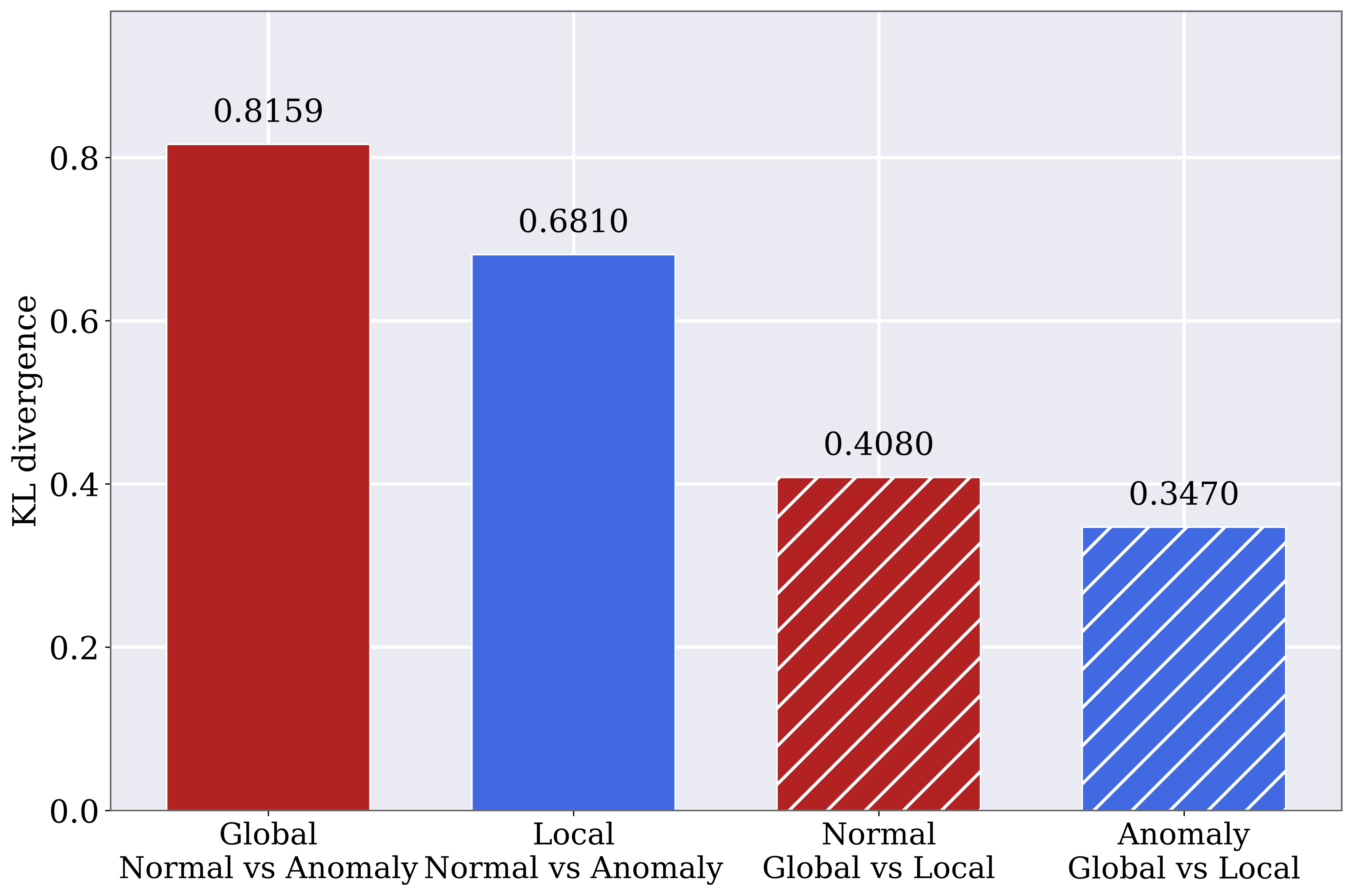}
        \caption{JS divergence of prompt embeddings }
    \end{subfigure}
    \caption{Quantitative analysis of distributional differences across global/local and normal/anomaly prompts using KL divergence, Wasserstein distance, and JS divergence.}
    \label{fig:Probability comparision}
\end{figure}

Fig.~\ref{fig:Probability comparision} quantitatively analyzes the relationships among the four prompt distributions learned on VisA~\citep{zou2022spot}: global normal, global anomaly, local normal, and local anomaly. We evaluate their pairwise differences using KL divergence~\citep{kullback1997information}, Wasserstein distance~\citep{gabriel2019computational}, and Jensen--Shannon (JS) divergence~\citep{deza2009encyclopedia}. For KL divergence, we report the average of the two directional divergences to obtain a symmetric pairwise measure.

The largest KL divergence is observed between the global normal and global anomaly distributions (0.8159), followed by the corresponding normal--anomaly difference at the local level (0.6810). By contrast, the global--local differences within the same semantic state are smaller. The Wasserstein and JS measurements show a consistent overall pattern, with normal--anomaly differences being larger than the corresponding global--local differences. These results are consistent with the intended hierarchical distributional structure, in which global and local prompts representing the same state remain relatively close while normal and anomalous distributions are separated within each level. Overall, the analysis provides quantitative evidence that the learned prompt distributions preserve cross-level relatedness while maintaining distinct normal and anomalous representations.

\subsection{Analysis of step-wise prompt transformation}
\begin{figure}[pos=htbp]
    \centering
    \begin{subfigure}{0.31\textwidth}
        \centering
        \includegraphics[width=\textwidth]{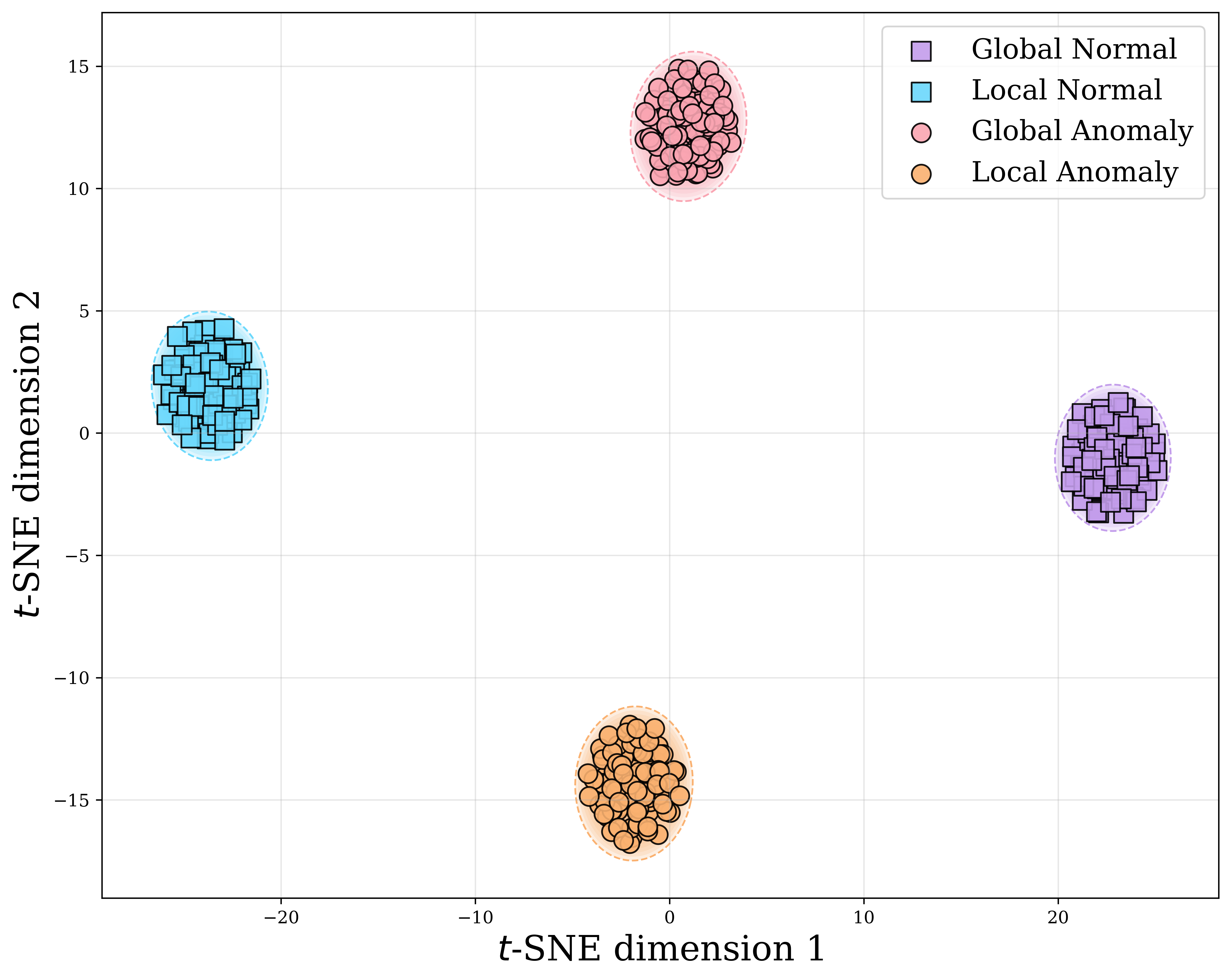}
        \caption{Distributional prompt space}
        \label{fig:t-sne(a)}
    \end{subfigure}
    % \hfill
    \begin{subfigure}{0.31\textwidth}
        \centering
        \includegraphics[width=\textwidth]{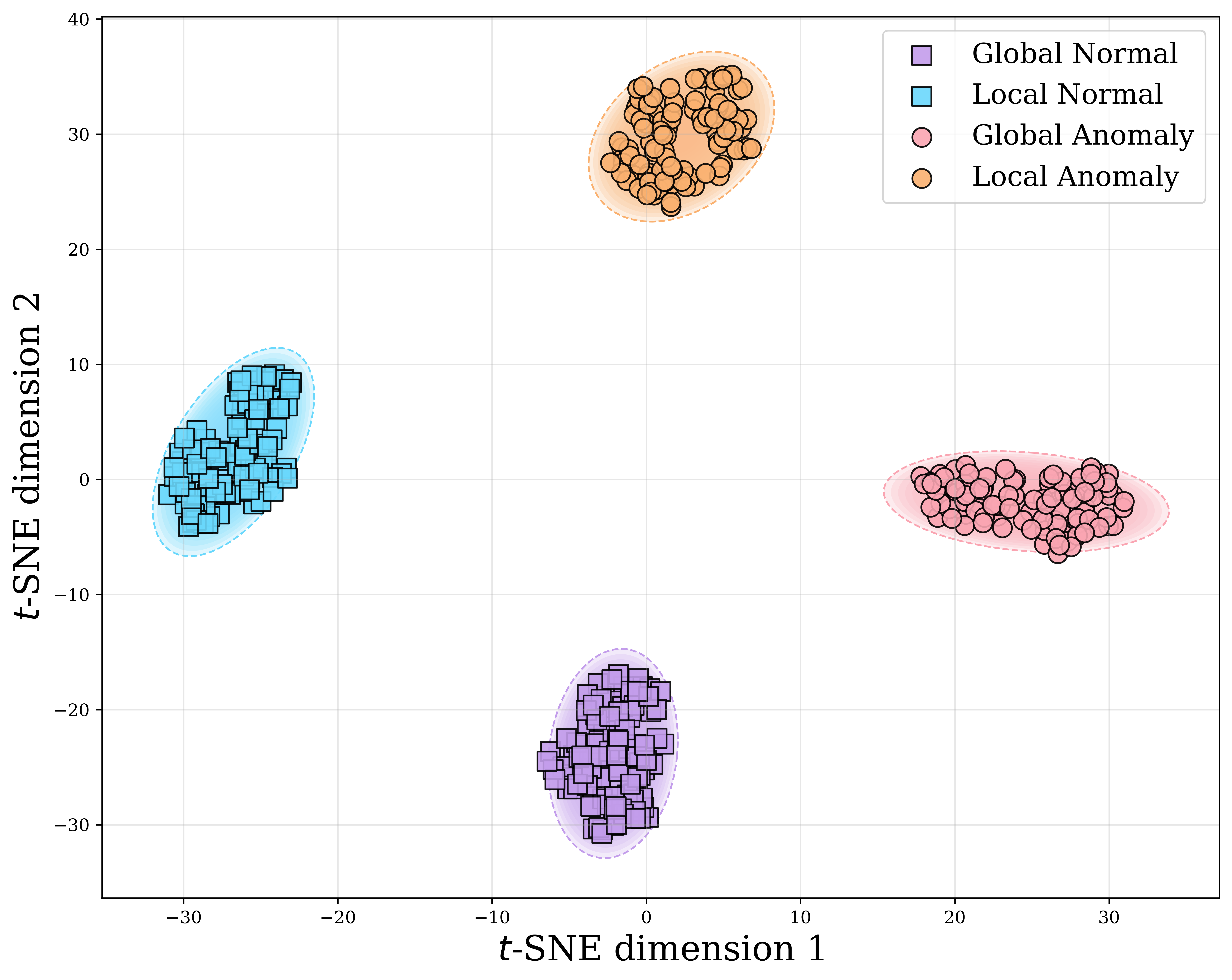}
        \caption{Textual embedding space}
        \label{fig:t-sne(b)}
    \end{subfigure}
    \begin{subfigure}{0.31\textwidth}
        \centering
        \includegraphics[width=\textwidth]{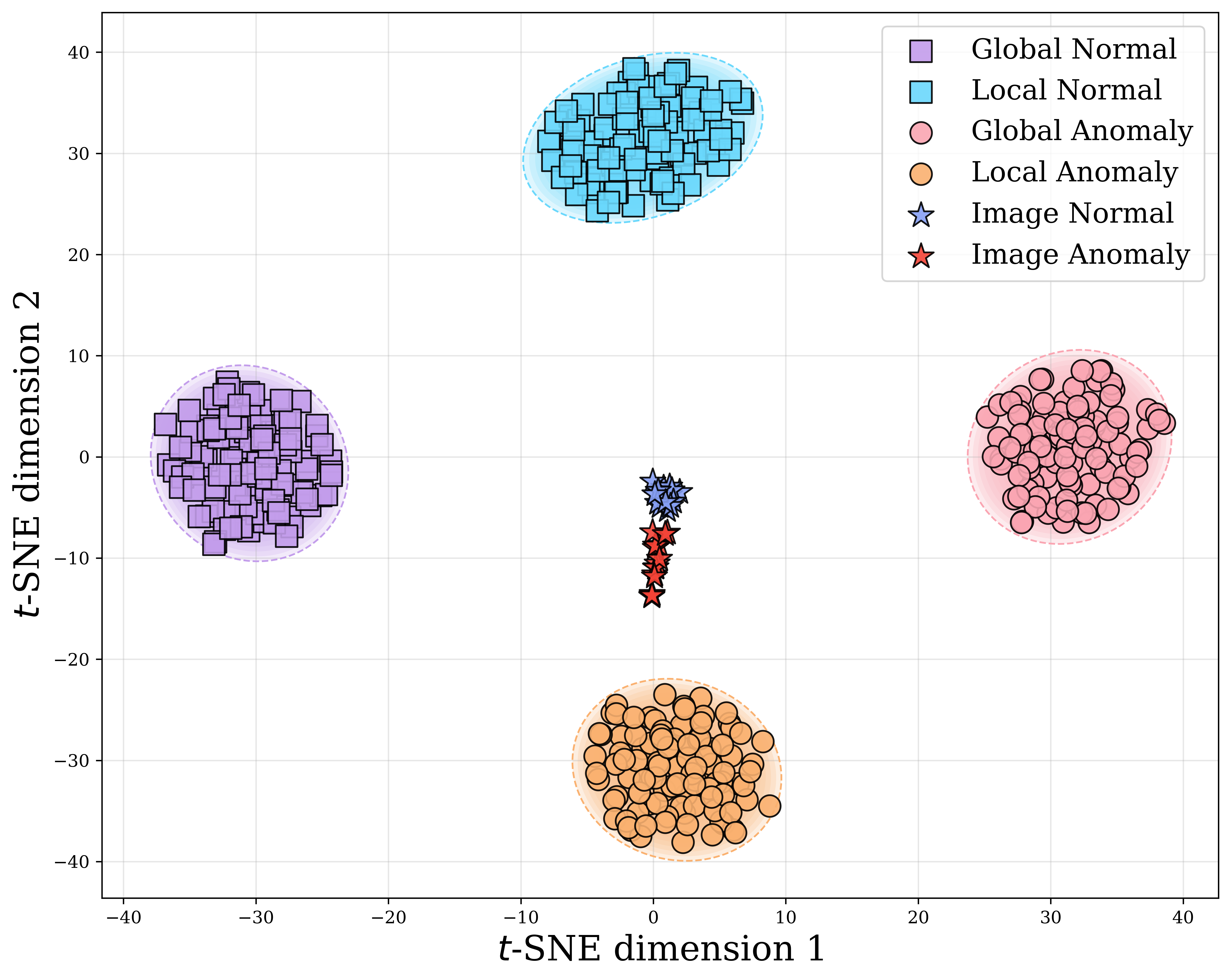}
        \caption{Textual-visual embedding space}
        \label{fig:t-sne(c)}
    \end{subfigure}
    \caption{\textit{t}-SNE visualizations of learned representation spaces.}
    \label{fig:t-sne}
\end{figure}

The \textit{t}-SNE visualizations in Fig.~\ref{fig:t-sne} illustrate how the learned prompt representations evolve across three stages: the distributional prompt space, the textual embedding space, and the final textual--visual embedding space.

Fig.~\ref{fig:t-sne}(a) visualizes the prompt distributions learned on VisA~\citep{zou2022spot} with $K=20$ samples. The global normal, global anomaly, local normal, and local anomaly prompts form distinguishable clusters, indicating that the learned distributions encode differentiated representations for the four prompt types. After text encoding, Fig.~\ref{fig:t-sne}(b) shows that these distinctions remain visible in the textual embedding space while the sampled prompts occupy broader regions within each group. This suggests that the text encoder preserves the relative separation among the learned prompt types while mapping their stochastic variations into the semantic embedding space.

Fig.~\ref{fig:t-sne}(c) further visualizes the joint textual--visual embedding space using images from the MVTec-AD~\citep{bergmann2019mvtec} hazelnut class. Normal prompt representations are located closer to normal image features, whereas anomalous prompt representations are more closely associated with anomalous image features. This pattern is consistent with the intended cross-modal alignment induced during training. Overall, the step-wise visualization provides qualitative evidence that the distinctions established in the learned prompt distributions are largely preserved through text encoding and subsequently organized with visual representations in the joint embedding space.

\subsection{Sensitivity to sampling size and prompt length}
\begin{figure}[pos=htbp]
    \centering
    \begin{subfigure}{0.33\textwidth}
        \centering
        \includegraphics[width=\textwidth]{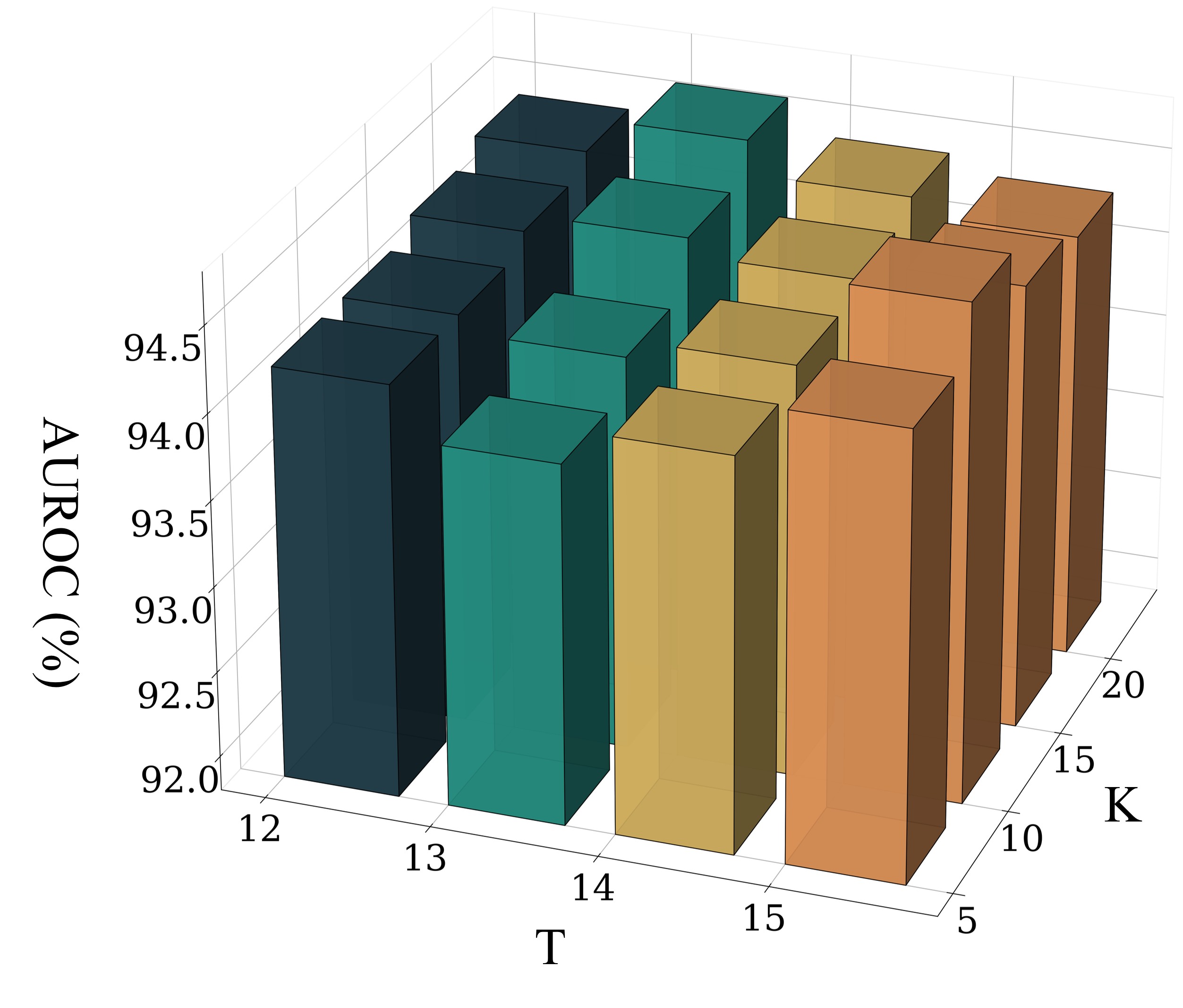}
        \caption{Image-level results}
    \end{subfigure}
    % \hfill
    \begin{subfigure}{0.33\textwidth}
        \centering
        \includegraphics[width=\textwidth]{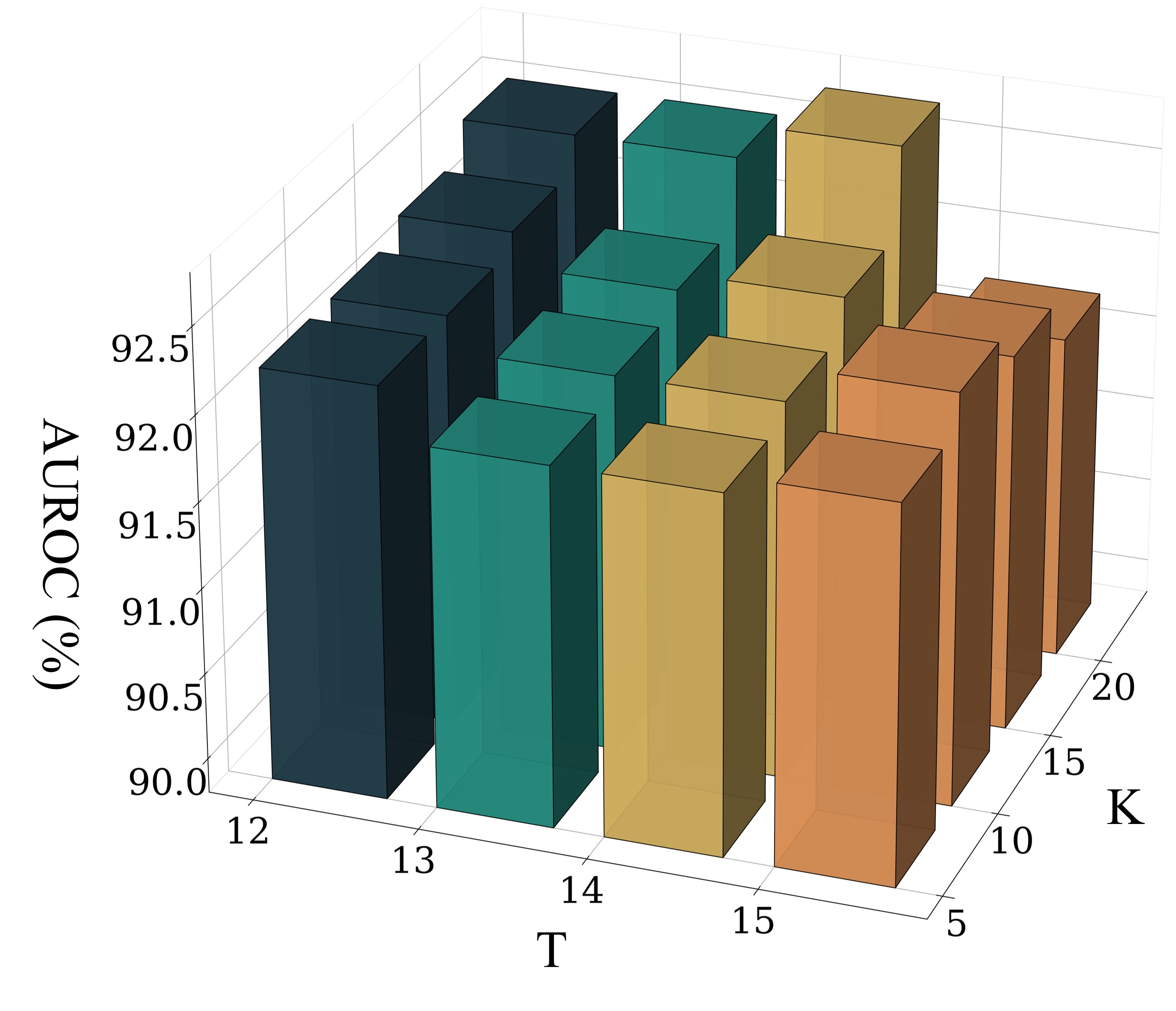}
        \caption{Pixel-level results}
    \end{subfigure}
    \caption{Comparison of image- and pixel-level performance under varying prompt lengths and $K$-sampling on MVTec-AD~\citep{bergmann2019mvtec}.}
    \label{fig:k-len-comparison}
\end{figure}

Fig.~\ref{fig:k-len-comparison} illustrates the effect of prompt length $T$ (12--15) and sampling size $K$ (5--20) on image- and pixel-level AUROC on MVTec-AD~\citep{bergmann2019mvtec}. For prompt lengths between 12 and 14, increasing $K$ generally improves both image- and pixel-level performance, suggesting that using more samples provides a more effective representation of the learned prompt distributions. However, increasing the prompt length to 15 does not yield further improvement and instead leads to a performance decrease. These results indicate that increasing either prompt length or sampling size does not monotonically improve performance, and that an appropriate combination of $T$ and $K$ is important for effective distributional prompt modeling.

% Conclusion and Future Work
\section{Conclusion}
We presented HiDist, a normality-referenced hierarchical distributional prompt framework for zero-shot anomaly detection. Rather than representing normal and anomalous states as independent point-estimate prompts, HiDist models prompt tokens distributionally and parameterizes anomalous distributions relative to learned normal references. These distributions are further specialized for global image-level detection and local pixel-level localization, while knowledge-guided hierarchical alignment using the adapted global visual representation and complementary regularization objectives organize the learned representations across semantic states and hierarchical levels.

Extensive experiments on industrial and medical benchmarks demonstrate strong cross-dataset performance of HiDist for both anomaly detection and localization. HiDist achieves competitive or superior results across most evaluated settings, while the ablation studies consistently support the contributions of normality-referenced distributional modeling, hierarchical representations, and the associated learning objectives. Further analyses of the learned distribution parameters and embedding spaces show that normal/anomalous and global/local prompts form differentiated yet structured representations, consistent with the intended design of HiDist.

Despite these results, several limitations remain. First, although multiple stochastic prompts are sampled from each learned distribution, their encoded representations are aggregated into a single prototype for subsequent inference. This aggregation reduces the number of text representations involved in online similarity computation from $K$ samples to a single representative prototype, but it also discards sample-level variability that could potentially provide complementary information during anomaly scoring. Second, the diagonal Gaussian formulation provides a compact parameterization of token-wise variations but imposes a unimodal and dimension-independent inductive bias, which may limit the representation of more complex prompt semantics. Future work may therefore explore aggregation mechanisms that preserve informative sample-level variations while retaining the compact inference of prototype-based scoring. Another direction is to develop more expressive distributional formulations beyond diagonal Gaussians while preserving the explicit normality-referenced dependency and global--local specialization central to HiDist.

\section*{CRediT authorship contribution statement}

\textbf{Jiyul Ham}: Writing -- original draft, Conceptualization, Methodology, Visualization, Validation, Writing -- review and editing;
\textbf{Seongho Yoon}: Validation, Visualization, Writing -- original draft,
Writing -- review and editing;
\textbf{Yonggon Jung}: Formal analysis, Validation, Writing -- review and editing;
\textbf{Sangjae Lee}: Investigation, Visualization, Writing -- review and editing;
\textbf{Jun-Geol Baek}: Investigation, Supervision, Writing -- review and editing.

\section*{Declaration of competing interest}
The authors declare that they have no known competing financial interests or personal relationships that could have appeared to influence the work reported in this paper.

\section*{Data availability}
Data will be made available on request.

\section*{Acknowledgements}
This work was supported by a National Research Foundation of Korea (NRF) grant funded by the Korean government (MSIT) (RS-2025-24534467, RS-2026-25520490), and the Institute of Information \& Communications Technology Planning \& Evaluation (IITP)-ICT Challenge and Advanced Network of HRD (ICAN)(RS-2024-00438411). This research was also supported by the BK21 FOUR funded by the Ministry of Education of Korea and National Research Foundation of Korea and Samsung Electronics Co., Ltd. (IO251218-14783-01).

\bibliographystyle{cas-model2-names}
\bibliography{references} 
\end{document}